\documentclass[journal]{IEEEtran}
\usepackage{booktabs}
\usepackage{graphicx}
\usepackage{subfig}
\usepackage{amssymb}
\usepackage{enumitem,amssymb}
\usepackage{bbding} 
\usepackage{cite}
\usepackage{indentfirst}
\usepackage{enumitem}
\usepackage{hyperref}
\usepackage{multirow}
\usepackage{makecell}  
\usepackage{nicematrix}
\usepackage[ruled]{algorithm2e}

\makeatletter
\newcommand{\removelatexerror}{\let\@latex@error\@gobble}
\makeatother
\makeatletter
\renewcommand{\maketag@@@}[1]{\hbox{\m@th\normalsize\normalfont#1}}%
\makeatother

\hypersetup{
	colorlinks=true,
	linkcolor=red,
	urlcolor=blue
}

\usepackage{tikz,xcolor}
\definecolor{lime}{HTML}{A6CE39}
\DeclareRobustCommand{\orcidicon}{%
	\begin{tikzpicture}
	\draw[lime, fill=lime] (0,0) 
	circle [radius=0.134] 
	node[white] {{\fontfamily{qag}\selectfont \tiny ID}};    \draw[white, fill=white] (-0.0625,0.095) 
	circle [radius=0.007];    \end{tikzpicture}
	\hspace{-2mm}}
\foreach \x in {A, ..., Z}{%
	\expandafter\xdef\csname orcid\x\endcsname{\noexpand\href{https://orcid.org/\csname orcidauthor\x\endcsname}{\noexpand\orcidicon}}
}

\begin{document}
	
\captionsetup[figure]{name={Fig.},labelsep=period}
\captionsetup[table]{name={TABLE},labelsep=period}

\title{Deep Blind Super-Resolution for Satellite Video}

\author{Yi~Xiao\orcidA{},
       Qiangqiang~Yuan\orcidB{}, ~\IEEEmembership{Member,~IEEE,}
       Qiang~Zhang\orcidD{}, 
       and~Liangpei~Zhang\orcidC{}, ~\IEEEmembership{Fellow,~IEEE.} 

\thanks{This work was supported in part by the National Natural Science Foundation of China under Grant 42230108 and Grant 61971319. (\emph{Corresponding author: Qiangqiang Yuan}.)}
\thanks{Yi Xiao is with the School of Geodesy and Geomatics, Wuhan University, Wuhan 430079, China (e-mail: xiao\_yi@whu.edu.cn).}
\thanks{Qiangqiang Yuan is with the School of Geodesy and Geomatics, Wuhan University, Wuhan 430079, China (e-mail: yqiang86@gmail.com).}
\thanks{Qiang Zhang is with the Information Science and Technology College, Dalian Maritime University, Dalian 116000, China (e-mail: qzhang95@dlmu.edu.cn).}
\thanks{Liangpei Zhang is with the State Key Laboratory of Information Engineering in Surveying, Mapping, and Remote Sensing, Wuhan University, Wuhan 430079, China (e-mail: zlp62@whu.edu.cn).}}

%
%

\markboth{IEEE Transactions on Geoscience and Remote Sensing}%
{Shell \MakeLowercase{\textit{et al.}}: Bare Demo of IEEEtran.cls for IEEE Journals}
%



\maketitle

\begin{abstract}
Recent efforts have witnessed remarkable progress in Satellite Video Super-Resolution (SVSR). However, most SVSR methods usually assume the degradation is fixed and known, \emph{e.g.}, \emph{bicubic} downsampling, which makes them vulnerable in real-world scenes with multiple and unknown degradations. To alleviate this issue, blind SR has thus become a research hotspot. Nevertheless, existing approaches are mainly engaged in blur kernel estimation while losing sight of another critical aspect for VSR tasks: temporal compensation, especially compensating for blurry and smooth pixels with vital sharpness from severely degraded satellite videos. Therefore, this paper proposes a practical Blind SVSR algorithm (BSVSR) to explore more sharp cues by considering the pixel-wise blur levels in a coarse-to-fine manner. Specifically, we employed multi-scale deformable convolution to coarsely aggregate the temporal redundancy into adjacent frames by window-slid progressive fusion. Then the adjacent features are finely merged into mid-feature using deformable attention, which measures the blur levels of pixels and assigns more weights to the informative pixels, thus inspiring the representation of sharpness. Moreover, we devise a pyramid spatial transformation module to adjust the solution space of sharp mid-feature, resulting in flexible feature adaptation in multi-level domains. Quantitative and qualitative evaluations on both simulated and real-world satellite videos demonstrate that our BSVSR performs favorably against state-of-the-art non-blind and blind SR models. 
Code will be available at \url{https://github.com/XY-boy/Blind-Satellite-VSR}
\end{abstract}

\begin{IEEEkeywords}
Satellite video, multiple degradations, blind super-resolution, deformable attention, remote sensing.
\end{IEEEkeywords}

%
\IEEEpeerreviewmaketitle

\section{Introduction}

\IEEEPARstart{V}{ideo} satellite has recently received increasing attention due to its strength in dynamic observations. Nowadays, satellite video imagery is widely used in remote sensing tasks with high temporal variations \cite{zhang2021_DPLRTSVD}, such as object tracking \cite{9672083}, anomaly detection \cite{guo2023anomaly, xu2022hyperspectral}, classification \cite{10041948, 9576700, li2019deep, 9763451,duan2023classification, he2022generating}, change detection \cite{luo2023multiscale}, \emph{etc}. However, the satellite platform tremors and atmosphere scatters in the remote imaging process often generate undesirable blurs in satellite videos. Additionally, the spatial resolution of satellite video is usually degraded to stabilize the remote transmission. As a result, the visual quality of satellite video is inevitably contaminated, which results in performance drops in subsequent applications. Therefore, it is of practical significance to improve the spatial resolution of satellite video for both human perception and downstream tasks.

\par Compared to hardware upgrades, Super-Resolution (SR) technology provides an optimal solution for this highly ill-posed problem \cite{arad2022ntire, yang2020mapping, jk4, xiao2022generating, he2023spectral}. Traditional SR methods \cite{5308275} often employ hand-crafted priors to make this problem well-posed. However, these methods are fragile to the laborious priors and may restore unsatisfactory results because the real constraint often deviates from the predefined priors. Besides, they also suffer from high computational complexity and can lead to sub-optimal performance.

\par Benefit from the development of deep neural networks (DNNs) \cite{wang2021estimating, wang2022spatiotemporal, 10143395}, especially the huge success in low-level vision tasks \cite{zhang2023_TNNLS, wang2022local, 9913829, jk3, jk5}, the deep-learning-based SR approaches are booming. In spite of achieving decent results, most of them are specialized for single and known degradation, \emph{e.g.}, \emph{bicubic} downsampling, and tend to collapse in real-scene. Therefore, more efforts have been paid to blind SR approaches, which address the SR problem under multiple and unknown degradations, \emph{e.g.}, unknown blur kernels, and downsamplings. Recently, numerous works have made promising progress on blind Single Image Super-Resolution (SISR). However, they are less applicable in satellite videos as the additional information in the temporal domain is not fully explored. More recently, some scholars have investigated blind Video Super-Resolution (VSR). Nevertheless, these methods focus solely on accurate blur kernel estimation while overlooking the significance of temporal compensation in VSR, especially precise compensation for blurry and smooth pixels in severely degraded frames.

\par Generally, existing blind VSR approaches typically utilize optical flow warping for temporal compensation. However, they are laborious in large-scale satellite imagery and not robust in complex imaging scenarios with scale variations and sparse motions. What's worse, in blind SR settings, the appearance of satellite video frames is severely blurred and downsampled, which poses more challenges for accurate flow estimation. Liu \emph{et al.} \cite{9745539} proposed a joint estimation network, which adopts patch matching to realize patch-by-patch alignment. However, the patch-wise alignment does not fully exploit the sub-pixel redundancy. Moreover, the high-similarity patch may also contain blurry pixels. Simply aggregating them to mid-feature is not sophisticated to consider the blur level of pixels, which may introduce and amplify the blurry information and restore unsatisfactory results. Overall, two issues hinder us from moving forward:
\begin{enumerate}
    \item Existing temporal compensation strategies are time-consuming and not robust in severely degraded satellite videos. 
    \item Lacking elaborate pixel-wise temporal modeling to grasp vital sharpness and eliminate unfavorable blurs.
\end{enumerate}

\par Although the pixels may be degraded by the same degree of blurring, they are still not equally informative to the recovery of the clean and HR. Therefore, we define sharp pixels as those that can provide more clean and sharp cues (\emph{e.g.}, high-frequency texture), which are beneficial for reconstruction. To address the issues mentioned above, a more practical manner is urgently needed to consider the blur level of pixels and explore more sharp and clean clues for temporal compensation. Inspired by previous research, Deformable Convolution (DConv) is a preferable choice in satellite videos as it benefits from adaptive pixel-wise sampling to mitigate the misalignment caused by inaccuracy optical flow estimation. However, the learned pixels in deformable sampling points are also blurry and not equally informative for restoring sharp details. Very recently, \cite{zhu2021deformable} has investigated the contribution of different sampling positions for efficient spatial element relationship modeling. This motivates us to model the different blurry levels of pixels with the equipment of deformable attention. Hence, we could effectively aggregate the deformable sampling points by encouraging the representation of sharp pixels and eliminating the effects of blurry pixels. 

\par  In particular, this paper proposes a novel approach (BSVSR) to progressively aggregate sharp information while considering the pixel-wise blur level. We adopt the efficient Multi-Scale Deformable convolution (MSD) alignment \cite{9530280} to explore multi-scale temporal redundancy from the entire frame sequence with window-slid progressive fusion. This helps to alleviate the alignment pressure brought by large displacements. To finely aggregate the sharp mid-feature, a multi-scale Deformable Attention (DA) module was proposed to measure the pixel-wise blur level, which is practical to assign more attention to clean and sharp pixels for better sharpness representation. Benefiting from the coarse-to-fine manner, we can favorably identify the critical sharpness from severely blurred and downsampled satellite videos. To flexibly adjust the solution space and make the sharp mid-feature adaptive to various degradation, a Pyramid Spatial Transform (PST) strategy is established. With pyramid design, we could improve the diversity of mid-feature with multi-level spatial activation, making the transformation aware of multi-scale spatial distribution in satellite videos.

\par To sum up, our contributions are listed as follows:

\begin{description}
\item 1) Different from previous optical-flow-based and patch-wise compensation methods, we propose to aggregate sharp information in severely degraded satellite videos with progressive temporal compensation, which exploits Multi-Scale Deformable (MSD) convolution and Deformable Attention (DA) to explore more sharp and clean clues by considering the blur level of pixels.
\item 2) To achieve flexible feature adaptation, we develop a robust Pyramid Spatial Transform (PST) module, where blur information could be transformed into mid-feature in multi-level feature domains.
\item 3) Extensive experiments are conducted on Jilin-1, Carbonite-2, UrtheCast, Skysat-1, and Zhuhai-1 video satellites. And the results demonstrate our BSVSR performs favorably against state-of-the-art blind and non-blind SR approaches. 
\end{description}

The remainder of this paper is organized as follows: Section \ref{related} reviews the progress of video super-resolution, Section \ref{meth} involves details of our approach, Section \ref{exp} includes extensive experiments and analysis, and Section \ref{conclu} is the conclusion.

\section{Related Work}\label{related}
\subsection{Deep-Learning-based Classical Super-Resolution}
We first review the classical SR as it lays the foundation of blind SR. Classical SR methods often assume that the degradation process is single and known, such as \emph{bicubic} downsampling. 

\subsubsection{Classical SISR methods}
With the success of SRCNN \cite{7115171}, CNN-based SISR methods have been blooming, with remarkable progress in deeper networks \cite{kim}, attention-based networks \cite{zhang2018image}, recurrent networks \cite{jk1} and recent popular transformer-based models \cite{liang2021swinir, he2022dster}. Although they achieved decent results in bicubic-downsampled images, they are not capable of handling multiple degradations. Also, lacking consideration of temporal redundancy makes SISR less generalized in VSR tasks.

\subsubsection{Classical VSR methods}
The key success of VSR tasks is to compensate the pixels of the mid-frame with the temporal redundancy along frames. According to the type of temporal compensation, classical VSR can be broadly divided into flow-based and kernel-based compensation. 
\par \textbf{Flow-based Compensation.} This approaches \cite{kappeler2016video, caballero2017real, haris2019recurrent} employs explicit optical flow to describe the motion relationships between frames and perform frame/feature-wise warping to align the adjacent frame to mid-frame. As mentioned in \cite{9745539}, optical flow estimation is time-consuming and not robust in blurry and low-resolution videos. Thus, the kernel-based method may provide a more efficient optimal for VSR task, benefiting from its adaptive learning capability.

\par \textbf{Kernel-based Compensation.} Such methods often implicitly involve the compensation into learnable parameters. Jo \emph{et al.} \cite{jo2018deep} proposed to directly learn a 3D upsampling filter (DUF) to super-resolve each LR pixel. 
To reduce the high computational consumption of 3D CNN, Tian \emph{et al.} \cite{tian2020tdan} introduce an efficient deformable network to explore more temporal priors with deformable sampling points. Later, more efforts have been paid to generating precise offset parameters for deformable sampling, such as pyramid \cite{wang2019edvr} and multi-scale \cite{9351768} architectures. Some works employ the attention mechanism to find out valuable complementary. Yu \emph{et al.} \cite{yu2022memory} established a novel cross-frame non-local attention and memory-augment attention (MANA) to memorize more details in mid-feature. Recently, some recurrent networks \cite{yi2021omniscient, chan2022basicvsr++} proposed to recurrently propagate temporal information for better compensation.

\par Although kernel-based approaches could realize adaptive compensation at the pixel level, they rarely consider the blur level of pixels and treat them equally, thus failing to find the most useful sharp cues which are beneficial for restoration. 

\subsection{Deep-Learning-based Blind Super-Resolution}
Depending on whether apparent blur information is transformed, these kinds of methods can be subdivided into explicit and implicit blur transformation. 

\par \textbf{Explicit Blur Transformation}. Early work \cite{Zhangkai} encoded the explicit blur kernels into feature maps and simply concatenated them with LR features. However, they only allow partial transformation. Later, \cite{usrnet} proposed a deep Unfolding Super-resolution Network (USRNet), which transforms the blur kernel by unfolding the optimization problem. To mitigate the performance drop caused by the mismatch between the estimated blur kernel and the realistic one,  \cite{gu2019blind} established an iterative kernel correct network (IKC) to adaptively refine the predicted blur kernel. In addition, they further proposed a Spatial Feature Transformation (SFT) layer to profoundly transform the blur information into features for the adaptive SR process. Particularly, the blur kernel is exploited to guide the shift and scaling of LR features, which helps the LR features to adapt to a suitable domain for reconstruction. Currently, the SFT layer has been commonly used for blur transformation. However, it still lacks flexibility in modulating the spatial and channel of deep features. Recently, Wang \emph{et al.} \cite{wang2021unsupervised} presented a Degradation-Aware Convolution (DAConv) to achieve deep blur transformation in two branches. In spatial branches, they use depth-wise convolution to integrate the blur representations into the LR feature. In the channel branch, channel attention was introduced for channel-wise modulation. 

\par \textbf{Implicit Blur Transformation}. Such methods mainly grasp implicitly learning the domain distribution from external datasets. Jo \emph{et al.} \cite{ada} implicitly learned an adaptive generator (AdaTarget) to upsample LR images with diverse unknown blur kernels. Similarly, some efforts \cite{bulat, 9022593, cingan} have been made to achieve cross-domain learning with Generative Adversarial Networks (GAN). However, the implicit transformation faces harsh convergence conditions as GAN-based models are easy to collapse and produce undesirable artifacts \cite{9870558}. 

In summary, the explicit blur transformation is mainstream because they are straightforward and easy to train. Nevertheless, adapting the feature to a desired domain still remains a challenging problem.

\subsection{Blind Super-Resolution for Satellite Video}
Most early works are SISR approaches \cite{wu2023lightweight, jk2, 9787539, 9107103, 8677274}. However, their performance has reached a plateau without exploring the temporal information. To grasp the temporal redundancy in satellite videos \cite{xiao2021recurrent}, Liu \emph{et al.} \cite{liu2020satellite} proposed a traditional VSR framework, which uses the non-local temporal similarity as priors to constrain the solution space. He \emph{et al.} \cite{he2021multiframe} employ 3D convolution to achieve temporal compensation, which is not elaborate in exploring the temporal priors. Xiao \emph{et al.} \cite{9530280} developed a multi-scale DConv for precise alignment and proposed a temporal grouping projection to fuse the aligned features. To explore the spatial-temporal collaborative redundancy, they put forward a flow-DConv deeply coupled strategy \cite{xiao2022space} and enhance the temporal information by a spatial-temporal transformer. Recently, Jin \emph{et al.} \cite{jin2023learning} introduced both transformer and CNN to fully excavate the local and global redundancy. Xiao \emph{et al.} \cite{xiao2023local} developed a novel framework that exploits temporal difference to realize temporal compensation. Although these methods perform favorably on bicubic-downsampled satellite videos, it is less generalized in blind degradation settings, \emph{e.g.}, unknown blurs and downsampling.

Towards this end, some blind super-resolution approaches \cite{xiao2023degrade, wu2022blind} have thus become a hot spot. Liu \emph{et al.} \cite{9745539} further proposed a joint estimation network for collaborative optimization of blur kernel estimation and SR process. He \emph{et al.} \cite{10003260} develop a ghost-module network for blind satellite VSR. However, they both lose sight of precise temporal compensation in severely blurry and low-resolution satellite videos. In fact, not all pixels are clean and sharp and beneficial for blind VSR. Therefore, we still need to move forward by compensating more clean and sharp cues for blurry and smooth pixels.

\begin{figure*}[!t]
\centering
\includegraphics[width=7in]{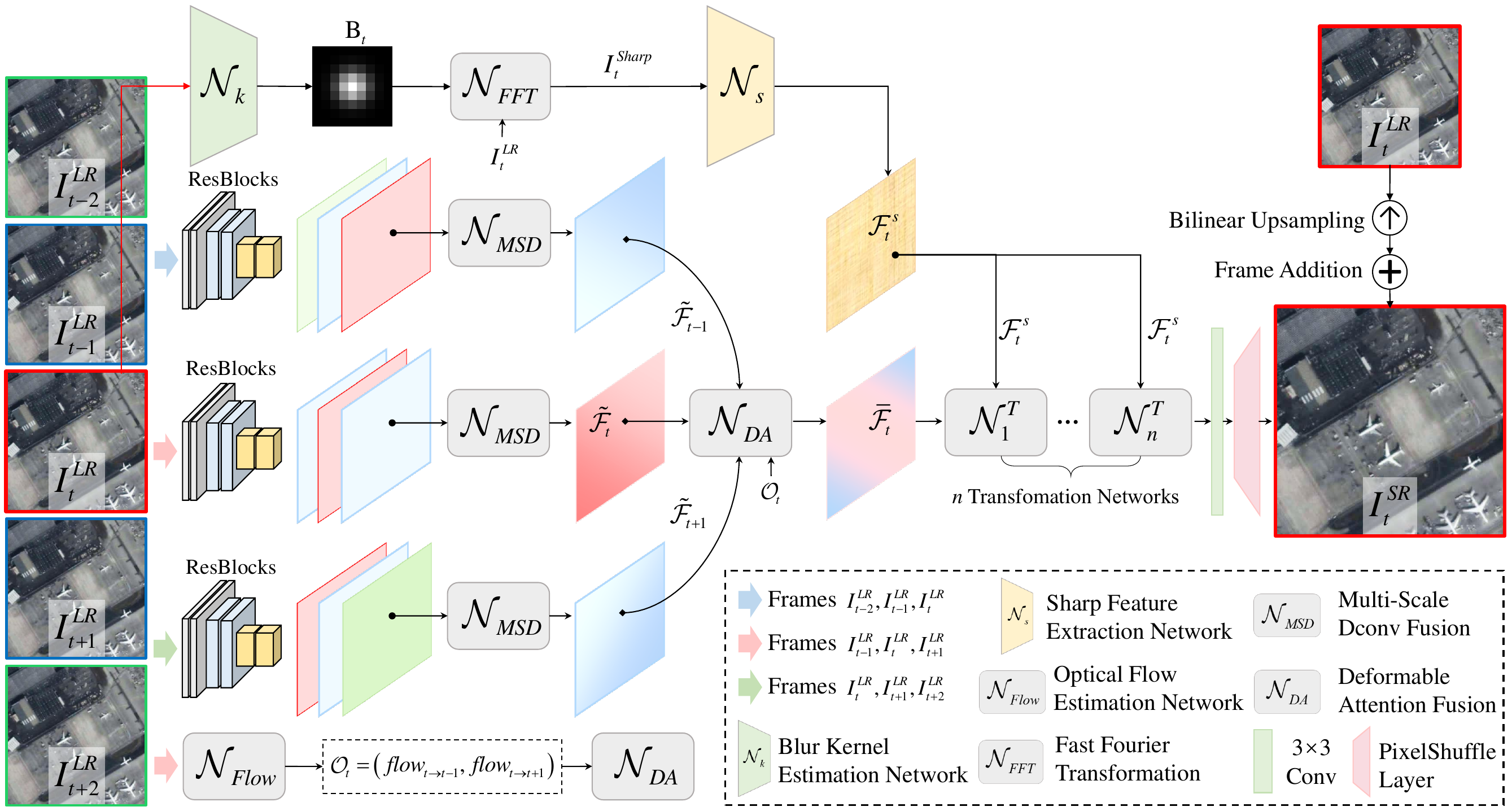}%
\captionsetup{font={scriptsize}}   
\caption{The overview of our BSVSR, which takes $2N+1=5$ consecutive blurry low-resolution frames as input and predicts the sharp high-resolution mid-frame $I^{SR}_t$.}
\label{fig1}
\end{figure*}

\begin{figure*}[ht]
\centering
\includegraphics[width=7in]{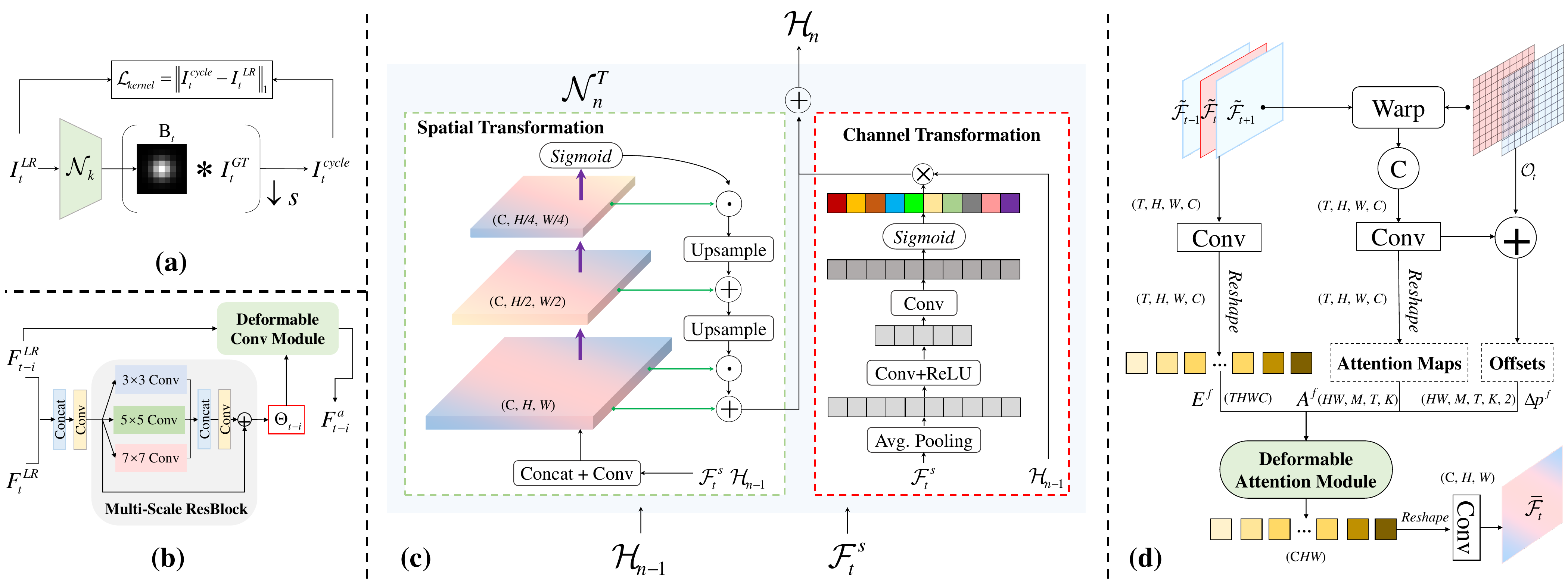}%
\captionsetup{font={scriptsize}}   
\caption{The flowchart of (a) the optimization of blur kernel estimation network; (b) Multi-Scale Deformable (MSD) convolution alignment \cite{9530280} used for coarse compensation; (c) n-th blur-aware transformation network, which receives blur-aware feature ${\rm{{\cal F}}}_t^s$ and the result of previous transformation ${{\rm{{\cal H}}}_{n-1}}$ and outputs ${{\rm{{\cal H}}}_{n}}$; (d) Deformable Attention (DA) used for fine compensation.}
\label{fig3}
\end{figure*}

\begin{figure*}[ht]
\centering
\includegraphics[width=7in]{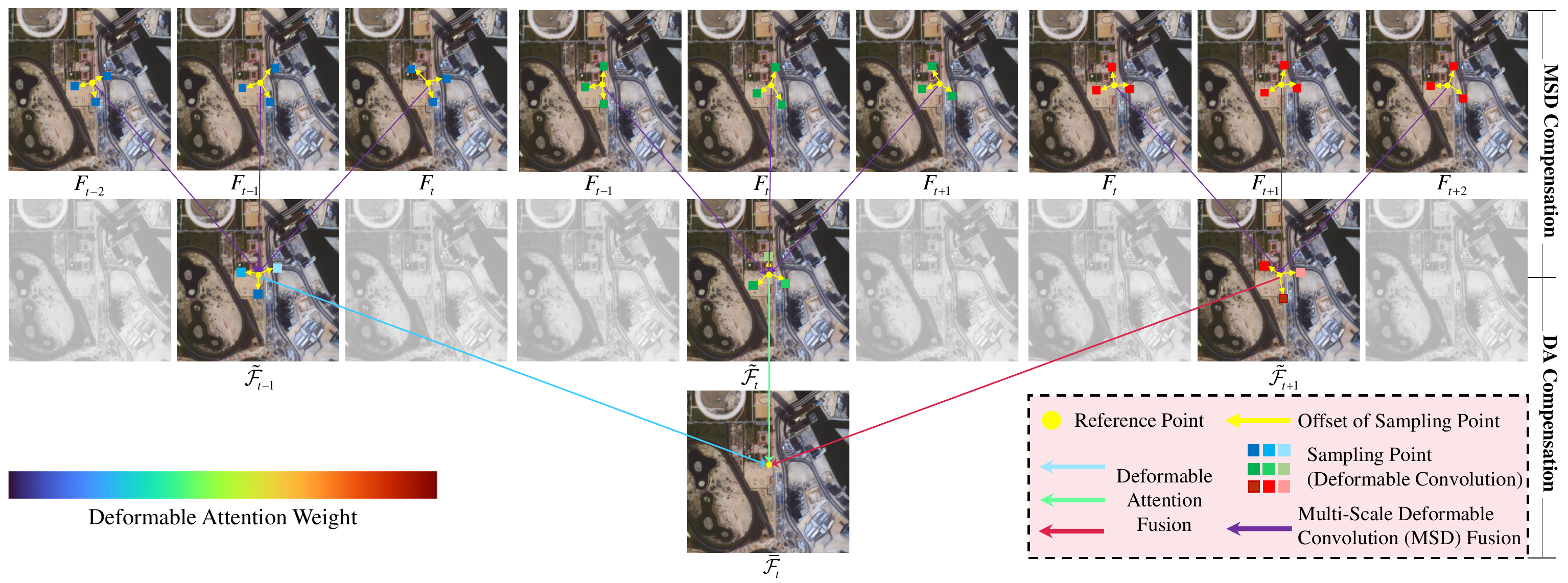}%
\captionsetup{font={scriptsize}}   
\caption{The illustration of the proposed progressive temporal compensation strategy, which used \textbf{M}ulti-\textbf{S}cale \textbf{D}eformable convolution (\textbf{MSD}) compensation and \textbf{D}eformable \textbf{A}ttention (\textbf{DA}) fusion to explore more clean and sharp cues in a coarse-to-fine manner. The color of the sampling points represents the attention weights used for aggregation. By assigning higher attention weights to clean and sharp points, we can encourage the representation of vital sharpness and eliminate blurry pixels.} 
\label{fig2}
\end{figure*}

\section{Methodology}\label{meth}
\subsection{Degradation Formulation}
 Let $I^{LR}_t$ be the degraded mid-frame. The degradation process of $I^{LR}_t$ can be formulated by the following:
\begin{equation}
I_t^{LR} = {\left( {I_t^{GT} \otimes B} \right)_{ \downarrow s}} + N,\label{equa1}
\end{equation}
where $I_t^{GT}$ represents the HR ground-truth mid-frame, $B$ is a 2D blur kernel filter, $\otimes$ means convolution, $\downarrow s$ denotes spatial downsampling operation with a scale factor of $s$ and $N$ is the noise term. Following previous works \cite{9745539}, this paper only focuses on blur and downsampling degradations, which means $N=0$. And we set $B$ to isotropic Gaussian blur kernel and $\downarrow s$ to $\times s$ bicubic downsampling.
\subsection{Overview}
The proposed network ${f_{{\mathop{\rm BSVSR}\nolimits} }}\left(  \cdot  \right)$ aims to recover a clean and sharp high-resolution (HR) mid-frame $I^{SR}_t$ given $2N+1$ consecutive blurry LR frames ${\rm I} = \left\{ {I_{t + n}^{LR}} \right\}_{n =  - N}^N$. That is:
\begin{equation}
I_t^{SR} = {f_{{\mathop{\rm BSVSR}\nolimits} }}\left( {\rm I} \right).
\end{equation}
\par As shown in Fig. \ref{fig1}, we set $N=2$ for a brief illustration. Firstly, we perform blur estimation to generate blur-aware feature ${\rm{{\cal F}}}_t^s$ through ${{\rm{{\cal N}}}_k}$, ${{\rm{{\cal N}}}_{FFT}}$ and ${{\rm{{\cal N}}}_s}$. Secondly, the blurry LR frames will be progressively compensated to sharp mid-feature ${\rm{{\cal F}}}_t^s$ by multi-scale deformable convolution ${{\rm{{\cal N}}}_{MSD}}$ and multi-scale deformable attention ${{\rm{{\cal N}}}_{DA}}$ in a coarse-to-fine manner. 
Thirdly, ${\rm{{\cal F}}}_t^s$ is transformed into ${{\rm{\bar {\cal F}}}_t}$ to guide the sharp mid-feature adaptive to a desired domain for reconstruction. Denote the output of $n$-th transformation ${\rm{{\cal N}}}_n^T$ as ${{\rm{{\cal H}}}_n}$. 

Finally, we set a $3\times3$ convolution and a pixel-shuffle layer to restore the super-resolved mid-frame from ${{\rm{{\cal H}}}_n}$.

In the following subsections, we will introduce the implementation details of our BSVSR.
\subsection{Blur Estimation}
We follow the blur estimation process in \cite{pan2021deep} as it offers decent performance with a lightweight design. Firstly, we employ kernel estimation network ${{\rm{{\cal N}}}_k}$ to predict the explict blur kernel ${{\mathop{\rm B}\nolimits} _t}$ from $I_t^{LR}$. 
\begin{equation}
{{\mathop{\rm B}\nolimits} _t} = {{\rm{{\cal N}}}_k}\left( {I_t^{LR}} \right).
\end{equation}
As shown in Fig. \ref{fig3} (a), the estimated blur kernel is used to generate $I^{cycle}_t$. ${{\rm{{\cal N}}}_k}$ is trained by optimalize the $L_1$ difference between $I^{cycle}_t$ and $I^{LR}_t$. After training of ${{\rm{{\cal N}}}_k}$, a Fast Fourier Transformation strategy ${{\rm{{\cal N}}}_{FFT}}$ is adopted to explore the latent sharp mid-frame $I_t^{sharp}$. 
\begin{equation}
I_t^{sharp} = {{\rm{{\cal N}}}_{FFT}}\left( {I_t^{LR},{{\mathop{\rm B}\nolimits} _t}} \right)
\end{equation}
Through a sharp feature extraction network ${{\rm{{\cal N}}}_s}$, we obtain a blur-aware mid-feature ${\rm{{\cal F}}}_t^s$ that contains blur information.
\begin{equation}
{\rm{{\cal F}}}_t^{s} = {{\rm{{\cal N}}}_s}\left( {I_t^{sharp}} \right)
\end{equation}

Note that blur estimation is not the major focus of our paper and we pay more attention on accurate compensation for blurry LR frames. More details about ${{\rm{{\cal N}}}_k}$, ${{\rm{{\cal N}}}_{FFT}}$ and ${{\rm{{\cal N}}}_s}$ can be found at \cite{pan2021deep}.

\subsection{Coarse-to-fine Progressive Compensation}
Previous blind VSR methods use optical flow warping to align adjacent frames to mid-frame for compensation, which tends to introduce inaccurate motions in severely blurred LR satellite frames. To mitigate the misalignment and extract sharp information, we proposed a progressive compensation strategy that aggregates LR features in a coarse-to-fine manner. As shown in Fig. \ref{fig2}, it consists of two stages, termed Multi-Scale Deformable convolution (MSD) compensation and Deformable Attention (DA) compensation. Before performing compensation, $2N+1$ consecutive frames ${\rm I}$ will be extracted to LR features $\left\{ {F _{t + n}^{LR}} \right\}_{n =  - N}^N$.
\par \textbf{Multi-Scale Deformable convolution (MSD) Compensation.} The MSD compensation layer employs multi-scale deformable convolution to align consecutive frames in three slid-windows ${W_i}\left( {i = t - 1,t,t + 1} \right)$, where ${W_i} = \left\{ {F_{i - 1}^{LR},F_i^{LR},F_{i + 1}^{LR}} \right\}$ contains three consecutive frames. For simplicity, we use $i=t$ as an example to explain how to compensate $F_t^{LR} $ with $F_{t - 1}^{LR}$. The key to DConv is learning additional offsets for sampling points. As shown in Fig.\ref{fig3} (b), the adopted MSD could generate precise offset by exploring multi-scale features in satellite videos. The offset is generated by:
\begin{equation}
{\Theta _{t - 1}} = {\mathop{\rm MSRB}\nolimits} \left( {{\mathop{\rm Conv}\nolimits} \left[ {F_t^{LR},F_{t - 1}^{LR}} \right]} \right) = \left\{ {\Delta {p_k}} \right\}_{k = 1}^K,
\end{equation}
where $\rm MSRB$ is multi-scale residual block, $\left[  \cdot  \right]$ is feature concatenation, $K$ means the sampling numbers. Here, $K=9$ represents 9 sampling points in a $3\times3$ convolution grid.
\par For a position $p_0$ in $F_{t-1}^{LR}$, the location of 9 sampling points is $\left( p_0 + p_k \right)$, where ${p_k} \in \left\{ {\left( { - 1, - 1} \right), \cdots ,\left( {0,0} \right), \cdots ,\left( {1,1} \right)} \right\}$. With the addition offset $\Delta {p_k}$, the deformable sampling location can be written as $\left( p_0 + p_k + \Delta {p_k} \right)$. Denote the weight of $k$-th sampling point as ${\omega _k}$, the value of compensated pixel at position $p_0$ can be formulated by:
\begin{equation}
F_{t - 1}^a\left( {{p_0}} \right) = \sum\nolimits_{k = 1}^K {{\omega _k} \cdot F_{t - 1}^{LR}\left( {{p_0} + {p_k} + \Delta {p_k}} \right)} 
\end{equation}
Similarly, we could obtain the compensated feature $F_{t}^a$ and $F_{t + 1}^a$. The coarse compensated feature ${{\rm{\tilde {\cal F}}}_t}$ is fused by a $3\times3$ convolution:
\begin{equation}
{{\rm{\tilde {\cal F}}}_t} = {\mathop{\rm Conv}\nolimits} \left( {\left[ {F_{t - 1}^a,F_t^a,F_{t + 1}^a} \right]} \right).
\end{equation}
By setting $i=t-1$ and $i=t+1$, we could get coarse compensated features ${{\rm{\tilde {\cal F}}}_{t-1}}$ and ${{\rm{\tilde {\cal F}}}_{t+1}}$ from windows $W_{t-1}$ and $W_{t+1}$ respectively.

\begin{table*}[ht]
  \centering
\captionsetup{font={scriptsize}}   
  \caption{The properties of five video satellites involved in this paper, including Jilin-1, UrtheCst, Carbonite-2, SkySat-1, and Zhuhai-1.}
    \begin{tabular}{ccccccc}
\toprule
    Train/Test & Video Satellite & Region & Captured Date & Duration (s) & Frame rate (f/ps) & Frame Size \\
\midrule
    \multirow{8}[0]{*}{Train} & \multirow{8}[0]{*}{Jilin-1} & San Francisco, USA & April 24th, 2017 & 20    & 25    & 3840 × 2160 \\
          &       & Derna, Libya & May 20th, 2017 & 30    & 25    & 4096 × 2160 \\
          &       & Valencia, Spain & May 20th, 2017 & 30    & 25    & 4096 × 2160 \\
          &       & Tunisia & May 25th, 2017 & 30    & 25    & 4096 × 2160 \\
          &       & Adana-02, Turkey & May 25th, 2017 & 30    & 25    & 4096 × 2160 \\
          &       & Minneapolis-01, USA & June 2nd, 2017 & 30    & 25    & 4096 × 2160 \\
          &       & Minneapolis-02, USA & June 2nd, 2017 & 30    & 25    & 4096 × 2160 \\
          &       & Muharag, Bahrain & June 4th, 2017 & 30    & 25    & 4096 × 2160 \\
\midrule
    \multirow{7}[0]{*}{Test} & Jilin-1 & San Diego, USA & May 22th, 2017 & 30    & 25    & 4096 × 2160 \\
          &   Jilin-1    & Adana-01, Turkey & May 25th, 2017 & 30    & 25    & 4096 × 2160 \\
          &  UrtheCast & Cape Town, South Africa & -    & 12    & 30    & 1920 × 1080 \\
          &   UrtheCast    & Vancouver, Canada & -     & 34    & 30    & 1920 × 1080 \\
          & Carbonite-2 & Mumbai, India & April, 2018 & 60    & 6     & 2560 × 1440 \\
          & Skysat-1 & Las Vegas, USA & March 25th, 2014 & 60    & 30    & 1920 × 1080 \\
          & Zhuhai-1 & Dalin, China & -    & 29    & 25    & 1920 × 1080 \\
\bottomrule
    \end{tabular}%
  \label{data}%
\end{table*}%

\par \textbf{Deformable Attention (DA) Compensation.}
The Deformable Attention compensation layer implicitly models the pixel-wise blur level to explore more sharp and clean cues for the final mid-feature.
\par As shown in Fig. \ref{fig3} (d), ${{\rm{{\cal N}}}_{DA}}$ receives ${{\rm{\tilde {\cal F}}}_{t-1}}$, ${{\rm{\tilde {\cal F}}}_{t}}$, ${{\rm{\tilde {\cal F}}}_{t+1}}$ and optical flows ${{{\rm{{\cal O}}}_t}}$ as input and produces ${{\rm{\bar {\cal F}}}_t}$, where ${{\rm{{\cal O}}}_t} = \left( {flo{w_{t \to t - 1}},flo{w_{t \to t + 1}}} \right)$ contains two optical maps from frame $I_t^{LR}$ to its neighboring frames $I_{t-1}^{LR}$ and $I_{t+1}^{LR}$. Here, we adopt PWC-Net \cite{sun2018pwc} to predict the optical flows as it offers decent performance. Notably, different from previous works that rely on optical flows for compensation \cite{pan2021deep, 10003260}, ${{{\rm{{\cal O}}}_t}}$ is mainly used to generate the base offsets to stabilize the optimization of deformable attention.
\par \textbf{Step 1.} We concatenated the coarse features to generate a tensor with the shape of $T\times H\times W\times C$, where $H$, $W$, $C$ denotes the height, width, and channel number of frame features, $T=3$ means the feature numbers. After a convolution layer, the tensor will be unfolded to a flattened feature ${E^f} \in {\mathbb{R}^{THWC}}$.
\begin{equation}
{E^f} = {\mathop{\rm ReShape}\nolimits} \left( {{\mathop{\rm Conv}\nolimits} \left( {\left[ {{{{\rm{\tilde {\cal F}}}}_{t - 1}},{{{\rm{\tilde {\cal F}}}}_t},{{{\rm{\tilde {\cal F}}}}_{t + 1}}} \right]} \right)} \right).
\end{equation}
\par \textbf{Step 2.} Before fusing coarse compensated features, they are back warped by optical flows. 
\begin{equation}
{\rm{{\cal F}}}_k^w = {\mathop{\rm Warp}\nolimits} \left( {{{{\rm{\tilde {\cal F}}}}_k},flo{w_{i \to k}}} \right),k = i - 1,i + 1,
\end{equation}
Then, we concatenated the warped features and pass them to a convolution layer to generate a tensor of shape $T\times H\times W\times C$. On the one hand, it will be reshaped to an attention map ${A^f} \in {\mathbb{R}^{HW \times M \times T \times K}}$. In this manner, the sharpness of pixels could be measured by the attention maps.
\begin{equation}
{A^f} = {\mathop{\rm ReShape}\nolimits} \left( {{\mathop{\rm Conv}\nolimits} \left( {\left[ {{\rm{{\cal F}}}_{i - 1}^w,{\rm{{\cal F}}}_{i + 1}^w} \right]} \right)} \right).
\end{equation}
where $M$ is the heads of multi-head self-attention, and $K=9$ represents the number of sampling points.
On the other hand, we add it with the optical flows to provide the base offset of sampling points $\Delta {P^f} \in {\mathbb{R}^{HW \times M \times T \times K \times 2}}$:
\begin{equation}
\Delta {P^f} = {\mathop{\rm ReShape}\nolimits} \left( {{\mathop{\rm Conv}\nolimits} \left( {\left[ {{\rm{{\cal F}}}_{i - 1}^w,{\rm{{\cal F}}}_{i + 1}^w,} \right]} \right) + {{\rm{{\cal O}}}_i}} \right).
\end{equation}
\par \textbf{Step 3.} The flatten feature $E^f$, base offsets of sampling points $\Delta {P^f}$ and attention maps $A^f$ are sent into deformable attention function $\rm{{\cal D}}$ to generate the compensated flatten feature of shape $C\times H\times W$. And finally, we reshape it and pass it to a convolution layer:
\begin{equation}
{{\rm{\bar {\cal F}}}_t} = {\mathop{\rm Conv}\nolimits} \left( {{\mathop{\rm ReShape}\nolimits} \left( {{\rm{{\cal D}}}\left( {{E^f},{A^f},\Delta {P^f}} \right)} \right)} \right).
\end{equation}
\par Note that the deformable attention function also introduces a multi-scale design to generate multi-level offsets, which quite fits the characteristic of remote sensing imagery with multi-scale objects. For more details of the deformable attention function, please refer to \cite{zhu2021deformable}.

\subsection{Blur-aware Transformation}
To adapt the sharp mid-feature into a suitable domain for restoration, we need to make the features aware of blur information by effective transformation.
\par As shown in Fig. \ref{fig3} (c), the transformation happens in two branches, \emph{i.e.} Pyramid Spatial Transformation and Channel Transformation. Take the $n$-th transformation network ${\rm{{\cal N}}}_n^T$ as an example. The output of ${\rm{{\cal N}}}_n^T$ is determined by:
\begin{equation}
{{\rm{{\cal H}}}_n} = {\rm{{\cal N}}}_n^T\left( {{\rm{{\cal F}}}_t^s,{{\rm{{\cal H}}}_{n - 1}}} \right),
\end{equation}
where ${{\rm{{\cal H}}}_n}$ is the summation of the outputs derived from spatial and channel transformation branches, respectively. We denote the outputs as $F_t^{ST}$ and $F_t^{CT}$. ${{\rm{{\cal H}}}_{n-1}}$ is the output of ${\rm{{\cal N}}}_{n-1}^T$. In the following, we will describe how to obtain the spatial and channel-transformed features.

\par Pyramid Spatial Transformation. This branch aims to find a robust spatial activation to modulate the features for better blur awareness. Firstly, we gained the transformed feature $L_1$ at the 1st level by concatenating and fusing ${\rm{{\cal N}}}_{n-1}^T$ and ${\rm{{\cal F}}}_t^s$:
\begin{equation}
{L_1} = {\mathop{\rm Conv}\nolimits} \left( {\left[ {{{\rm{{\cal H}}}_{n - 1}},{\rm{{\cal F}}}_t^s} \right]} \right).
\end{equation}
Through two bilinear downsampling layers, the pyramid features $L_2$ at the 2nd and $L_3$ at the 3rd levels can be obtained. Such pyramid design helps to preserve the multi-scale information in satellite imagery \cite{fang2023hyperspectral}, thereby leading to accurate spatial activation. The spatial activation can be formulated by:
\begin{equation}
att_s = {\mathop{\rm UP}\nolimits} \left( {{\mathop{\rm UP}\nolimits} \left( {\sigma \left( {{L_3}} \right) \odot {L_3}} \right) + {L_2}} \right),
\end{equation}
With the help of this spatial activation, we could perform blur-transformation in the spatial dimension, which means the spatial transformed feature can be obtained by:
\begin{equation}
F_t^{ST} = att_s \odot {L_1} + {L_1}.
\end{equation}

\par \textbf{Channel Transformation.} We exploit the widely used Sequence-and-Excitation Network (SENet) \cite{hu2018squeeze} as the channel transformation algorithm, given its efficient calculation and decent performance. The channel-wise modulation weight $at{t_c}$ is generated by the following:
\begin{equation}
at{t_c} = \sigma \left( {{\mathop{\rm Squeeze}\nolimits} \left( {{\rm{{\cal F}}}_t^s} \right)} \right),
\end{equation}
where squeeze operation ${\mathop{\rm Squeeze}\nolimits} \left(  \cdot  \right)$ includes an average pooling layer, a convolution layer with ReLU activation for feature compression, and another convolution for feature expanding. Here $\sigma$ is \emph{sigmoid} function for activation. And finally, the channel transformed feature $F_t^{CT}$ can be determined by:
\begin{equation}
F_t^{CT} = at{t_c} \odot {{\rm{{\cal F}}}_t^s},
\end{equation}
where $\odot$ represents the channel-wise multiplication.

\subsection{Feature Extraction and Reconstruction}
\textbf{Feature Extraction.} The proposed BSVSR uses five residual blocks to extract $2N+1$ blurry LR features $\left\{ {F_{t + n}^{LR} \in {\mathbb{R}^{HWC}}} \right\}_{n =  - N}^N$, where $C$ is set to 128 in our final model. Each residual block consists of "Conv+ReLU+Conv" and is equipped with a global shortcut.
\par \textbf{Reconstruction.} We set a $3\times3$ convolution ${\mathop{\rm Conv}\nolimits} \left(  \cdot  \right)$ and Pixel-shuffle layer ${\mathop{\rm PS}\nolimits} \left(  \cdot  \right)$ \cite{espcn} to restore the super-resolved mid-frame, which means:
\begin{equation}
I_t^{SR} = {\mathop{\rm PS}\nolimits} \left( {{\mathop{\rm Conv}\nolimits} \left( {{{\rm{{\cal H}}}_{n}}} \right)} \right) + {\left( {I_t^{LR}} \right)_{ \uparrow s}},
\end{equation}
where ${\left(  \cdot  \right)_{ \uparrow s}}$ denotes the bilinear upsampling with a scale factor of $s$.

\begin{table*}[!t]
  \centering
\captionsetup{font={scriptsize}} 
  \caption{Quantitative results on the Jilin-1 testset. Here, we evaluate the comparative methods under various blur kernel widths ($\sigma=1.2, 1.6, 2.0$). The best PSNR/SSIM performances are highlighted in bold \textcolor{red}{\textbf{red}} color, and the second-best results are marked in \textcolor{blue}{blue} color.}
  \renewcommand{\arraystretch}{1.2} 
  \linespread{1.0}

\begin{tabular}{c|c|c|ccccccc}
\Xhline{1.5pt}
\rule{0pt}{9pt}
$\sigma$ & Type & Method & Scene-1   & Scene-2   & Scene-3   & Scene-4   & Scene-5   & Scene-6   & Average \\
\hline
\rule{0pt}{9pt}
    \multirow{10}[8]{*}{1.2} & \multicolumn{1}{c|}{\multirow{2}[1]{*}{\makecell[c]{Non-blind \\ SISR}}} & Bicubic & 24.02/0.7992 & 23.54/0.7595 & 25.28/0.8304 & 23.59/0.7715 & 22.55/0.7228 & 22.77/0.7492 & 23.63/0.7722 \\
          &       & SwinIR \cite{liang2021swinir}& 30.49/0.9182 & 28.11/0.8767 & 29.40/0.9058 & 28.40/0.8861 & 27.11/0.8624 & 29.76/0.9096 & 28.88/0.8931 \\
\cline{2-10} \rule{0pt}{9pt}         & \multicolumn{1}{c|}{\multirow{3}[1]{*}{\makecell[c]{Blind \\ SISR}}} & IKC \cite{gu2019blind}   & 33.19/0.9434 & 30.86/0.9191 & 33.79/0.9475 & 31.84/0.9272 & 29.67/0.9042 & \textcolor{blue}{32.83}/0.9418 & 32.03/0.9305 \\
          &       & AdaTarget \cite{ada} & 30.38/0.9138 & 27.25/0.8615 & 30.03/0.9117 & 27.9/0.8787 & 25.55/0.8305 & 29.67/0.9132 & 28.46/0.8849 \\
          &       & DASR \cite{wang2021unsupervised}  & \textcolor{blue}{33.42}/0.9455 & \textcolor{red}{\textbf{31.43}}/0.9241 & \textcolor{blue}{34.18/0.9513} & \textcolor{blue}{32.29/0.9322} & \textcolor{red}{\textbf{30.36}}/\textcolor{blue}{0.9124} & 32.80/0.9414 & \textcolor{blue}{32.41/0.9345} \\
\cline{2-10} \rule{0pt}{9pt}         & \multicolumn{1}{c|}{\multirow{5}[1]{*}{\makecell*[c]{Non-blind \\ VSR}}} & DUF-52L \cite{jo2018deep} & 29.28/0.9160 & 29.00/0.8997 & 29.72/0.9190 & 28.81/0.9025 & 27.46/0.8798 & 29.31/0.9124 & 28.93/0.9049 \\
          &       & EDVR-L \cite{wang2019edvr} & 31.72/0.9394 & 30.41/0.9204 & 32.86/0.9457 & 27.76/0.8887 & 28.42/0.8939 & 30.69/0.9278 & 30.31/0.9193 \\
          &       & BasicVSR \cite{chan2021basicvsr} & 30.89/0.9241 & 29.02/0.8965 & 31.32/0.9268 & 30.09/0.9097 & 27.89/0.8801 & 29.90/0.9174 & 29.85/0.9091 \\
          &       & MSDTGP \cite{9530280} & 29.69/0.9173 & 30.45/0.9237 & 26.09/0.8581 & 29.29/0.9075 & 27.78/0.8929 & 32.62/\textcolor{blue}{0.9441} & 29.32/0.9073 \\
          &       & MANA \cite{yu2022memory}  & 32.14/0.9391 & 30.27/0.9201 & 29.18/0.9058 & 30.94/0.9236 & 27.36/0.8764 & 29.01/0.9143 & 29.82/0.9132 \\
\cline{2-10}\rule{0pt}{9pt}          & \multicolumn{1}{c|}{\multirow{2}[1]{*}{\makecell*[c]{Blind \\ VSR}}} & DBVSR \cite{pan2021deep} & 33.04/\textcolor{blue}{0.9459} & 30.80/\textcolor{blue}{0.9242} & 33.82/0.9512 & 31.87/0.9314 & 29.39/0.9072 & 32.52/0.9419 & 31.91/0.9336 \\
          &       & \textbf{BSVSR (ours)}  & \textcolor{red}{\textbf{33.98/0.9535}} & \textcolor{blue}{31.41}/\textcolor{red}{\textbf{0.9340}} & \textcolor[rgb]{ 1,  0,  0}{\textbf{34.87/0.9577}} & \textcolor[rgb]{ 1,  0,  0}{\textbf{32.54/0.9395}} & \textcolor{blue}{30.09}/\textcolor{red}{\textbf{0.9195}} & \textcolor{red}{\textbf{32.96/0.9469}} & \textcolor[rgb]{ 1,  0,  0}{\textbf{32.64/0.9418}} \\
    \hline
    \hline
    \rule{0pt}{9pt}
    \multirow{10}[8]{*}{1.6} & \multicolumn{1}{c|}{\multirow{2}[1]{*}{\makecell[c]{Non-blind \\ SISR}}} & Bicubic & 24.06/0.7952 & 23.57/0.7516 & 25.35/0.8267 & 23.65/0.7671 & 22.61/0.7165 & 22.79/0.7423 & 23.67/0.7666 \\
          &       & SwinIR \cite{liang2021swinir}& 30.20/0.9137 & 27.58/0.8650 & 28.95/0.8974 & 28.15/0.8790 & 27.06/0.8587 & 29.79/0.9090 & 28.62/0.8871 \\
\cline{2-10}\rule{0pt}{9pt}          & \multicolumn{1}{c|}{\multirow{3}[1]{*}{\makecell[c]{Blind \\ SISR}}} & IKC \cite{gu2019blind}   & 33.00/0.9415 & 30.66/0.9157 & 33.44/0.9447 & 31.67/0.9248 & 29.51/0.9009 & 32.78/0.9407 & 31.85/0.9287 \\
          &       & AdaTarget \cite{ada} & 30.93/0.9194 & 27.64/0.8672 & 29.45/0.9063 & 28.09/0.8811 & 25.97/0.8373 & 30.06/0.9154 & 28.69/0.8878 \\
          &       & DASR \cite{wang2021unsupervised}  & 33.05/0.9426 & 30.82/0.9183 & 33.78/0.9490 & 32.15/0.9302 & 29.88/0.9073 & 32.62/0.9398 & 32.05/0.9312 \\
\cline{2-10}\rule{0pt}{9pt}          & \multicolumn{1}{c|}{\multirow{5}[1]{*}{\makecell[c]{Non-blind \\ VSR}}} & DUF-52L \cite{jo2018deep} & 28.81/0.9057 & 28.36/0.8831 & 29.38/0.9107 & 28.46/0.8909 & 26.76/0.8603 & 28.60/0.8969 & 28.40/0.8913 \\
          &       & EDVR-L \cite{wang2019edvr} & 32.89/0.9446 & 30.79/0.9233 & 32.27/0.9411 & 27.51/0.8841 & 28.98/0.8991 & 30.39/0.9277 & 30.47/0.9200 \\
          &       & BasicVSR \cite{chan2021basicvsr} & 30.98/0.9240 & 28.76/0.8909 & 31.24/0.9251 & 29.96/0.9062 & 27.78/0.8749 & 29.67/0.9138 & 29.73/0.9058 \\
          &       & MSDTGP \cite{9530280} & 30.53/0.9259 & 31.02/\textcolor{blue}{0.9273} & 26.16/0.8602 & 30.51/0.91985 & 29.48/0.9110 & 32.78/\textcolor{blue}{0.9454} & 30.08/0.9149 \\
          &       & MANA \cite{yu2022memory}  & 32.33/0.9391 & 30.03/0.9155 & 28.23/0.8983 & 30.47/0.9188 & 26.81/0.8659 & 29.14/0.9147 & 29.50/0.9087 \\
\cline{2-10}\rule{0pt}{9pt}          & \multicolumn{1}{c|}{\multirow{2}[1]{*}{\makecell[c]{Blind\\ VSR}}} & DBVSR \cite{pan2021deep} & \textcolor{blue}{33.46/0.9474} & \textcolor{blue}{31.23}/0.9270 & \textcolor{blue}{34.23/0.9523} & \textcolor{blue}{32.32/0.9341} & \textcolor{blue}{29.95/0.9130} & \textcolor{blue}{32.87}/0.9441 & \textcolor{blue}{32.34/0.9363} \\
          &       & \textbf{BSVSR (ours)}  & \textcolor[rgb]{ 1,  0,  0}{\textbf{34.37/0.9548}} & \textcolor[rgb]{ 1,  0,  0}{\textbf{31.72/0.9354}} & \textcolor[rgb]{ 1,  0,  0}{\textbf{35.09/0.9579}} & \textcolor[rgb]{ 1,  0,  0}{\textbf{32.98/0.9419}} & \textcolor[rgb]{ 1,  0,  0}{\textbf{30.53/0.9230}} & \textcolor[rgb]{ 1,  0,  0}{\textbf{33.28/0.9485}} & \textcolor{red}{\textbf{33.00/0.9436}} \\
    \hline
    \hline
\rule{0pt}{9pt}
    \multirow{10}[8]{*}{2.0} & \multicolumn{1}{c|}{\multirow{2}[1]{*}{\makecell[c]{Non-blind \\ SISR}}} & Bicubic & 23.97/0.7862 & 23.45/0.7375 & 25.28/0.8187 & 23.58/0.7570 & 22.52/0.7035 & 22.66/0.7281 & 23.57/0.7552 \\
          &       & SwinIR \cite{liang2021swinir}& 29.25/0.9009 & 26.98/0.8508 & 28.15/0.8824 & 27.30/0.8615 & 26.66/0.8479 & 29.59/0.9056 & 28.00/0.8749 \\
\cline{2-10}\rule{0pt}{9pt}          & \multicolumn{1}{c|}{\multirow{3}[1]{*}{\makecell[c]{Blind \\ SISR}}} & IKC \cite{gu2019blind}   & 32.48/0.9364 & 30.42/0.9108 & 32.68/0.9378 & 31.25/0.9189 & 29.31/0.8962 & 32.54/0.9373 & 31.45/0.9229 \\
          &       & AdaTarget \cite{ada} & 30.81/0.9187 & 27.49/0.8615 & 28.62/0.8942 & 28.09/0.8786 & 26.15/0.8372 & 29.90/0.9102 & 28.51/0.8834 \\
          &       & DASR \cite{wang2021unsupervised}  & 32.86/0.9404 & 30.75/0.9161 & 33.45/0.9458 & 32.05/0.9283 & 29.85/0.9056 & 32.44/0.9374 & 31.90/0.9289 \\
\cline{2-10}\rule{0pt}{9pt}          & \multicolumn{1}{c|}{\multirow{5}[1]{*}{\makecell[c]{Non-blind \\ VSR}}} & DUF-52L \cite{jo2018deep} & 28.03/0.8877 & 27.43/0.8570 & 28.83/0.8967 & 27.76/0.8707 & 25.90/0.8311 & 27.61/0.8722 & 27.60/0.8692 \\
          &       & EDVR-L \cite{wang2019edvr} & 33.32/0.9454 & \textcolor{blue}{31.30/0.9291} & 32.92/0.9441 & 27.81/0.8871 & 29.74/0.9082 & 30.50/0.9297 & 30.93/0.9239 \\
          &       & BasicVSR \cite{chan2021basicvsr} & 30.46/0.9172 & 28.15/0.8784 & 30.54/0.9143 & 29.36/0.8949 & 27.39/0.8630 & 29.07/0.9027 & 29.16/0.8951 \\
          &       & MSDTGP \cite{9530280} & 30.37/0.9233 & 31.15/0.9265 & 26.21/0.8599 & 30.55/0.9206 & 29.94/\textcolor{blue}{0.9133} & \textcolor{blue}{32.99/0.9450} & 30.20/0.9148 \\
          &       & MANA \cite{yu2022memory}  & 31.40/0.9308 & 29.27/0.9027 & 26.86/0.8784 & 29.05/0.8981 & 26.14/0.8518 & 29.02/0.9097 & 28.63/0.8953 \\
\cline{2-10}\rule{0pt}{9pt}          & \multicolumn{1}{c|}{\multirow{2}[1]{*}{\makecell[c]{Blind\\ VSR}}} & DBVSR \cite{pan2021deep} & \textcolor{blue}{33.53/0.9463} & 31.30/0.9252 & \textcolor{blue}{34.06/0.9499} & \textcolor{blue}{32.35/0.9327} & \textcolor{blue}{30.03}/0.9117 & 32.81/0.9415 & \textcolor{blue}{32.35/0.9346} \\
          &       & \textbf{BSVSR (ours)}  & \textcolor[rgb]{ 1,  0,  0}{\textbf{34.58/0.9550}} & \textcolor{red}{\textbf{31.96/0.9362}} & \textcolor[rgb]{ 1,  0,  0}{\textbf{35.22/0.9578}} & \textcolor[rgb]{ 1,  0,  0}{\textbf{33.26/0.9431}} & \textcolor[rgb]{ 1,  0,  0}{\textbf{30.78/0.9241}} & \textcolor[rgb]{ 1,  0,  0}{\textbf{33.43/0.9490}} & \textcolor{red}{\textbf{33.21/0.9442}} \\
\Xhline{1.5pt}
    \end{tabular}%
  \label{table1}%
\end{table*}%

\begin{figure*}[!t]
\centering
\includegraphics[width=7in]{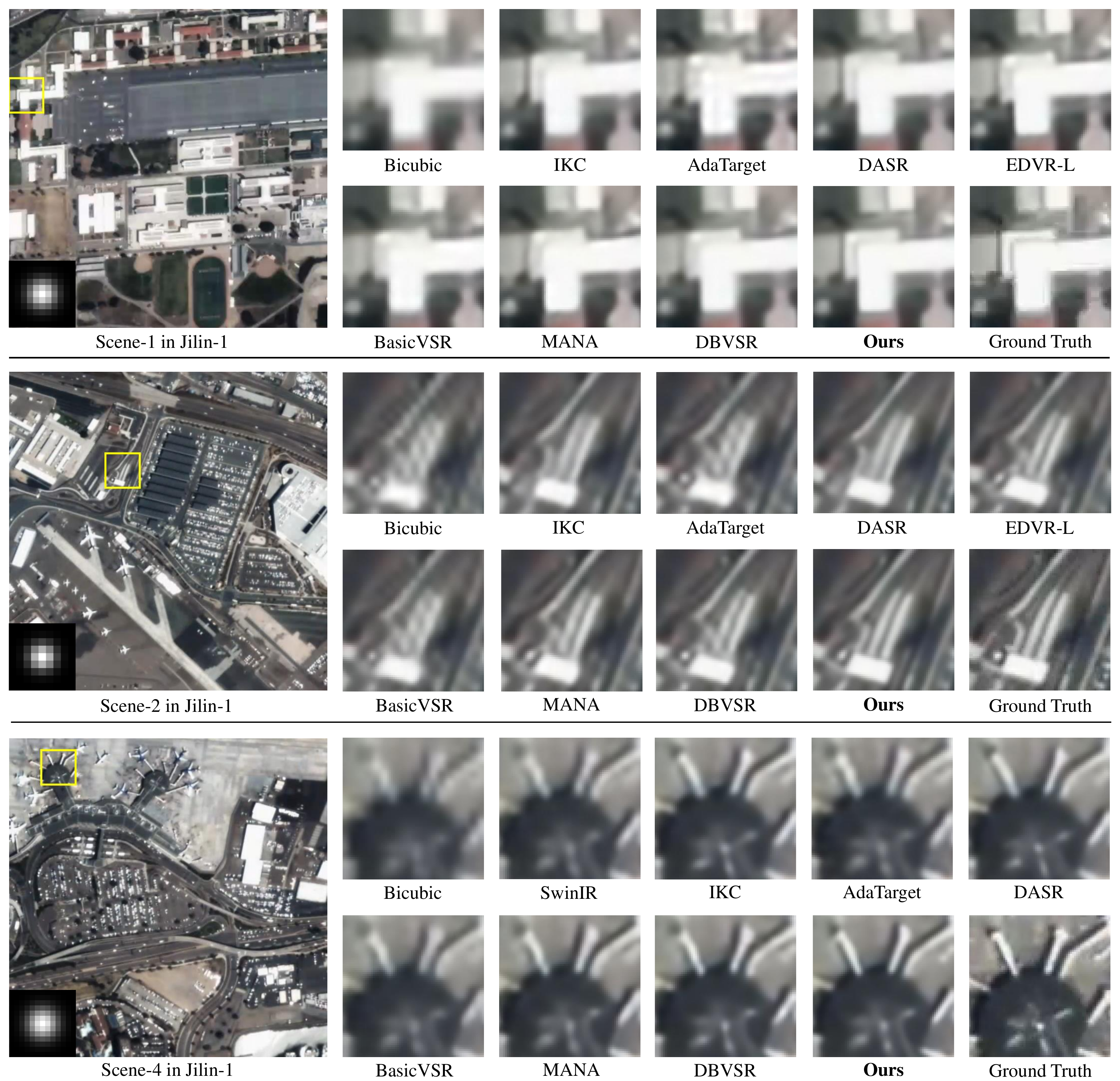}%
\captionsetup{font={scriptsize}}   
\caption{Qualitative results on Scene-1, Scene-2 and Scene-4 from Jilin-1 testset with various blur kernels. The size of the Region Of Interest (ROI) is $70 \times 70$. Our method recovers more sharp and clean details than state-of-the-art non-blind and blind SR methods.}
\label{jilin}
\end{figure*}

\begin{figure*}[!t]
\centering
\includegraphics[width=7in]{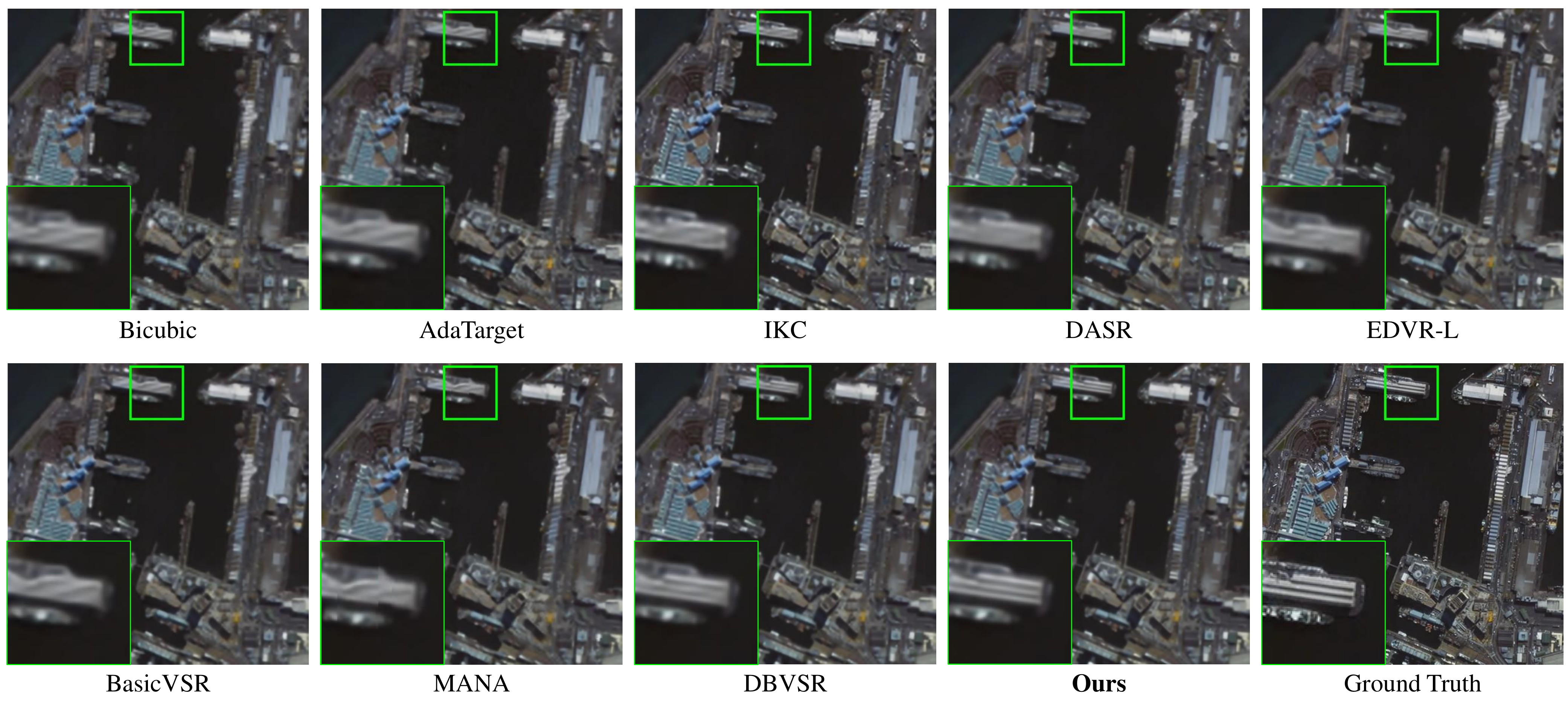}%
\captionsetup{font={scriptsize}}   
\caption{Qualitative results on UrtheCast test set with blur kernel width $\sigma=1.6$. The size of the Region Of Interest (ROI) is $90 \times 90$. Our method has fewer distortions and restores more textures than other state-of-the-art non-blind and blind SR methods.}
\label{uc}
\end{figure*}

\begin{figure*}[!t]
\centering
\includegraphics[width=7in]{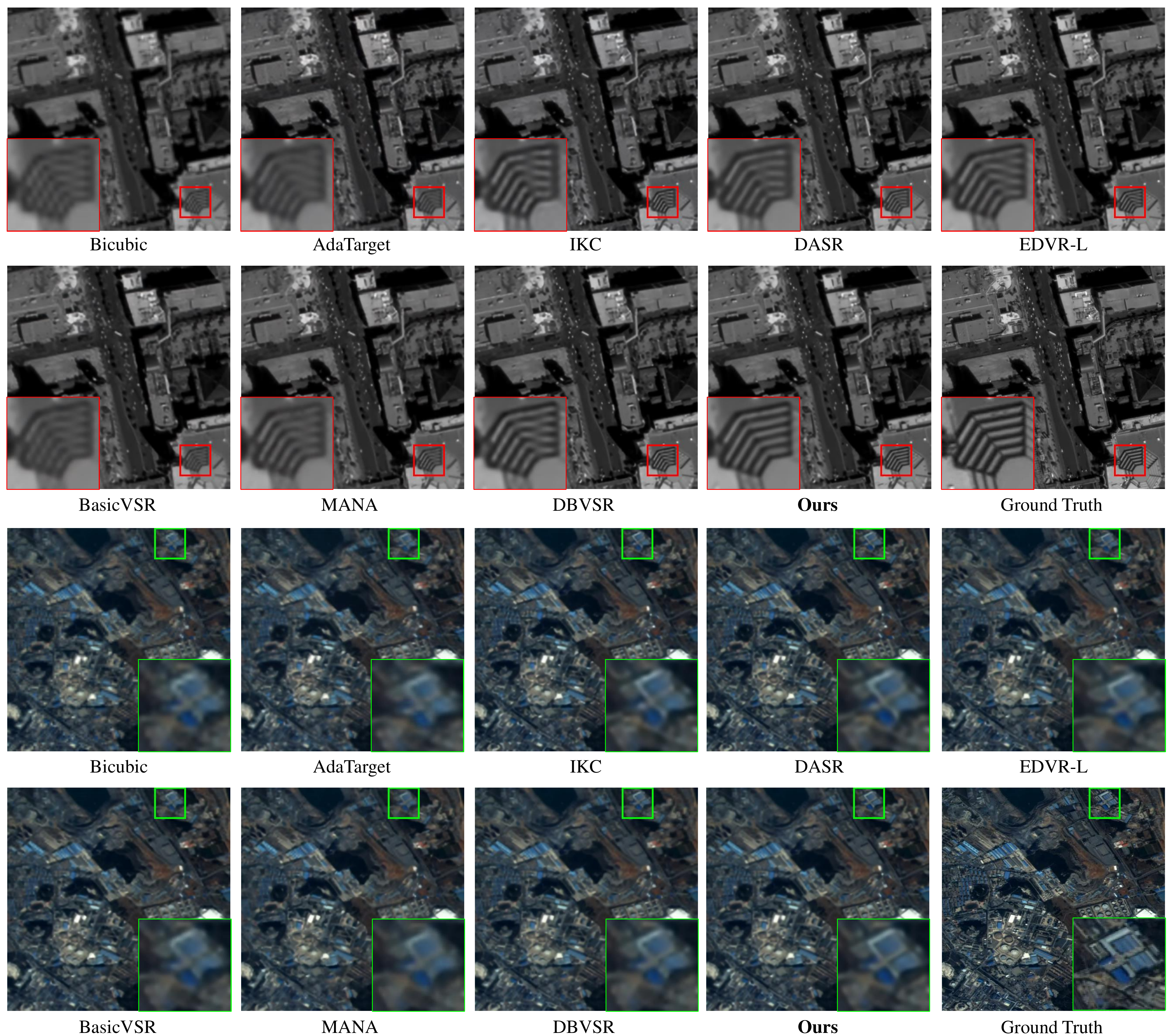}%
\captionsetup{font={scriptsize}}   
\caption{Qualitative results on SkySat-1 (top) and Zhuhai-1 (bottom) testsets with blur kernel width $\sigma=2.0$. The size of the Region Of Interest (ROI) is $70 \times 70$. Our method produces more realistic and sharp details than state-of-the-art non-blind and blind SR methods.}
\label{sky}
\end{figure*}

\begin{figure}[!t]
\centering
\includegraphics[width=3.5in]{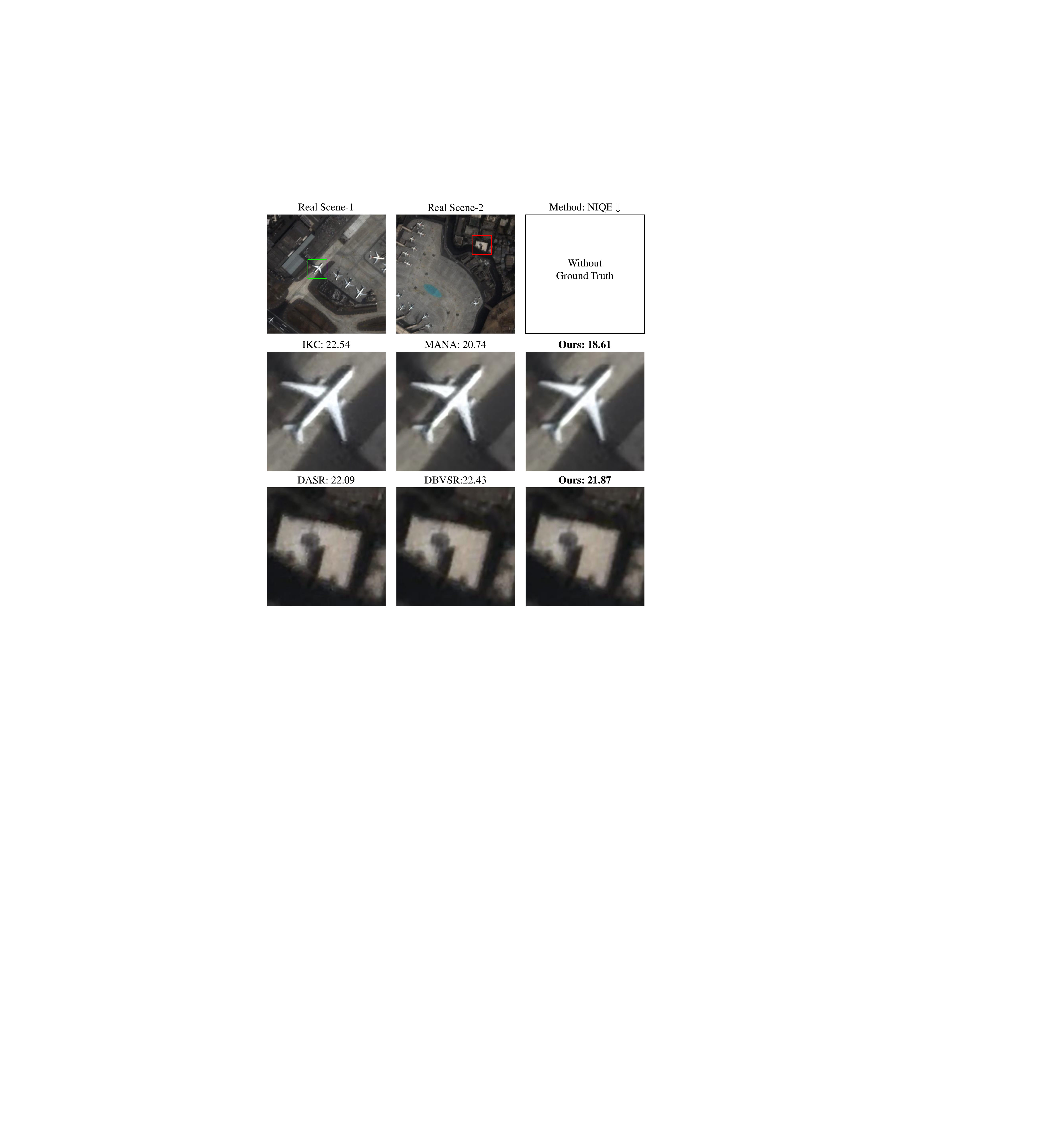}%
\captionsetup{font={scriptsize}}   
\caption{Real-world comparisons on Scene-1 and Scene-2 from Carbonite-2. Our method gains less blur on the boundary of the plane and building. A low NIQE represents a higher perceptual quality.}
\label{real}
\end{figure}

\begin{table*}[htbp]
  \centering
  \captionsetup{font={scriptsize}}
  \caption{Quantitative results on the Carbonite-2 and UrtheCast testsets. Here, the PSNR/SSIM is the average result of all scenes on each test set. The best PSNR/SSIM performances are highlighted in bold \textcolor{red}{\textbf{red}} color, and the second-best results are marked in \textcolor{blue}{blue} color. Note that BasicVSR is a recurrent method, and we followed the official settings, using 15 consecutive frames for propagation.}
    \begin{tabular}{c|cc|ccc|ccc}
\Xhline{1.5pt}
\rule{0pt}{9pt}
    \multirow{2}[1]{*}{Methods} & \multirow{2}[1]{*}{Input Frames} & \multirow{2}[1]{*}{Publication} & \multicolumn{3}{c|}{Carbonite-2} & \multicolumn{3}{c}{UrtheCast} \\
\cline{4-9} \rule{0pt}{9pt}          &       &       &   $\sigma=1.2$ & $\sigma=1.6$   & $\sigma=2.0$     & $\sigma=1.2$   & $\sigma=1.6$   & $\sigma=2.0$ \\
\hline
\rule{0pt}{9pt}
    Bicubic & -     & -     & 29.97/0.8445 & 30.04/0.8440 & 29.96/0.8395 & 25.77/0.7374 & 25.80/0.7315 & 25.70/0.7211 \\
    SwinIR \cite{liang2021swinir} & Single  & \emph{ICCV'21} & 30.74/0.8575 & 30.29/0.8484 & 30.29/0.8484 & 26.37/0.7589 & 26.09/0.7465 & 25.78/0.7328 \\
\hline
\rule{0pt}{9pt}
    IKC \cite{gu2019blind}   & Single   & \emph{CVPR'19} & \textcolor{red}{\textbf{36.97}}/0.9353 & \textcolor{blue}{36.74/0.9327} & \textcolor{blue}{36.48/0.9286} & \textcolor{red}{\textbf{29.98/0.8573}} & 29.72/\textcolor{blue}{0.8427} & \textcolor{blue}{29.65/0.8408} \\
    AdaTarget \cite{ada} & Single  & \emph{CVPR'21} & 32.17/0.8849 & 32.11/0.8861 & 31.90/0.8842 & 24.76/0.6900 & 24.90/0.6940 & 25.00/0.6960 \\
    DASR \cite{wang2021unsupervised}  & Single   & \emph{CVPR'21} & \textcolor{blue}{36.91/0.9358} & 36.36/0.9304 & 35.86/0.9245 & 29.84/0.8527 & \textcolor{blue}{29.81}/0.8408 & 29.54/0.8343 \\
\hline
\rule{0pt}{9pt}
    DUF-52L \cite{jo2018deep} & 7   & \emph{CVPR'18} & 32.00/0.8855 & 31.46/0.8769 & 31.05/0.8663 & 26.61/0.7661 & 26.32/0.7483 & 26.13/0.7348 \\
    EDVR-L \cite{wang2019edvr} & 5  & \emph{CVPRW'19} & 31.10/0.8635 & 32.07/0.8769 & 32.93/0.8898 & 24.87/0.7116 & 25.69/0.7444 & 26.55/0.7744 \\
    BasicVSR \cite{chan2021basicvsr} & Recurrent   & \emph{CVPR'21} & 33.62/0.9043 & 33.25/0.8983 & 32.63/0.8880 & 28.25/0.8212 & 28.08/0.8142 & 27.78/0.8055 \\
    MSDTGP \cite{9530280} & 5  & \emph{TGRS'22} & 29.17/0.8252 & 29.08/0.8264 & 28.88/0.8227 & 26.59/0.7564 & 26.34/0.7408 & 25.82/0.7214 \\
    MANA \cite{yu2022memory}  & 7  & \emph{CVPR'22} & 34.27/0.9174 & 33.42/0.9082 & 32.54/0.8946 & 25.72/0.7581 & 25.54/0.7475 & 25.48/0.7427 \\
\hline
\rule{0pt}{9pt}
    DBVSR \cite{pan2021deep} & 5 & \emph{ICCV'21} & 35.91/0.9280 & 35.92/0.9277 & 36.15/0.9289 & 29.64/0.8438 & 29.41/0.8351 & 29.02/0.8243 \\ 
 \textbf{BSVSR (ours)}  & 5  & - & 36.78/\textcolor{red}{\textbf{0.9359}} & \textcolor[rgb]{ 1,  0,  0}{\textbf{37.05/0.9378}} & \textcolor[rgb]{ 1,  0,  0}{\textbf{37.47/0.9401}} & \textcolor{blue}{29.90/0.8545} & \textcolor[rgb]{ 1,  0,  0}{\textbf{29.96/0.8500}} & \textcolor[rgb]{ 1,  0,  0}{\textbf{29.94/0.8453}} \\
\Xhline{1.5pt}
    \end{tabular}%
  \label{tab2}%
\end{table*}%

\begin{table*}[htbp]
  \centering
  \captionsetup{font={scriptsize}}
  \caption{Quantitative results on the SkySat-1 and Zhuhai-1 testsets. The best PSNR/SSIM performances are highlighted in bold \textcolor{red}{\textbf{red}} color, and the second-best results are marked in \textcolor{blue}{blue} color.}
    \begin{tabular}{c|c|cccccccc}
\Xhline{1.5pt}
\rule{0pt}{9pt}
    Test Set & $\sigma$ & Bicubic & IKC \cite{gu2019blind}   & DASR \cite{wang2021unsupervised}  & EDVR-L \cite{wang2019edvr} & BasicVSR \cite{chan2021basicvsr} & MANA \cite{yu2022memory} & DBVSR \cite{pan2021deep} & \textbf{BSVSR (ours)} \\
\hline
\rule{0pt}{9pt}
    \multirow{3}[0]{*}{SkySat-1} & 1.2   & 25.46/0.7573 & 31.36/0.8949 & \textcolor{red}{\textbf{31.81}}/0.9054 & 27.33/0.8375 & 29.87/0.8723 & 29.72/0.8841 & 31.63/\textcolor{blue}{0.9085} &  \textcolor{blue}{31.77}/\textcolor{red}{\textbf{0.9114}} \\
          & 1.6   & 25.50/0.7524 & 31.11/0.8897 & 31.53/0.8997 & 28.40/0.8593 & 29.75/0.8664 & 29.19/0.8735 & \textcolor{blue}{31.82/0.9087} &  \textcolor{red}{\textbf{31.97/0.9125}} \\
          & 2.0     & 25.40/0.7417 & 30.84/0.8831 & 31.29/0.8940 & 29.43/0.8746 & 29.13/0.8492 & 28.75/0.8600 & \textcolor{blue}{31.68/0.9041} &  \textcolor{red}{\textbf{32.06/0.9121}} \\
\hline
\rule{0pt}{9pt}
    \multirow{3}[0]{*}{Zhuhai-1} & 1.2   & 24.98/0.6390 &  \textcolor{red}{\textbf{28.06/0.7526}} &  \textcolor{blue}{27.91}/0.7501 & 25.03/0.6461 & 27.12/0.7242 & 25.74/0.6838 & 27.72/0.7468 & 27.79/\textcolor{blue}{0.7509} \\
          & 1.6   & 25.02/0.6300 &  \textcolor{blue}{27.59/0.7335} & 27.45/0.7298 & 25.20/0.6493 & 27.02/0.7136 & 25.37/0.6624 & 27.37/0.7235 & \textcolor{red}{\textbf{27.73/0.7338}} \\
          & 2.0     & 24.93/0.6160 & 27.43/0.7275 & 27.26/0.7169 & 25.92/0.6764 & 26.87/0.7036 & 25.03/0.6437 & \textcolor{blue}{27.52/0.7291} &  \textcolor{red}{\textbf{27.70/0.7311}} \\
\Xhline{1.5pt}
    \end{tabular}%
  \label{tab3}%
\end{table*}%

\section{Experiment and Discussion}\label{exp}
\subsection{Satellite Video Data source}
To evaluate the performance of our BSVSR against state-of-the-art approaches, we conduct comprehensive experiments on five mainstream video satellites, including Jilin-1, Carbonite-2, UrtheCast, SkySat-1, and Zhuhai-1. The training set is cropped from Jilin-1 satellite videos, where the spatial resolution of the ground-truth video is $640\times640$. Based on Equation \ref{equa1}, we degraded the high-resolution videos to generate low-resolution counterparts to establish the high-resolution and low-resolution training pairs. Following our previous works \cite{9530280, xiao2023local}, we ultimately obtained 189 video clips for model training. As for model testing, we randomly cropped 6 scenes from Jilin-1 that are non-overlapping with the training set. In addition, all the video clips of Carbonite-2, UrtheCast, SkySat-1, and Zhuhai-1 are used for further testing. The number of test video clips is 10, 14, 6, and 3. In the end, we had 39 videos from five video satellites for comprehensive model evaluation. A detailed summary of these five satellite videos can be found in Table \ref{data}. Our dataset is publicly available at \url{https://github.com/XY-boy}.

\subsection{Implementation Details}
We set $s=4$ to focus on $\times4$ blind SR for satellite video. As shown in Fig. \ref{fig1}, our network receives $2N+1=5$ consecutive frames as input, and the residual blocks extract homogeneous features with $C=128$ channels. For profound blur-aware transformation, we stack $n=20$ transformation module ${\rm{{\cal N}}}_n^T$, which strikes a balance between model size and performance. Consistent with previous blind VSR settings, we adopt isotropic Gaussian blur kernel of size $13\times13$ to degrade the ground-truth videos, as isotropic Gaussian blur kernel is highly consistent with the blur distribution in satellite videos\cite{9745539}. The standard deviation of Gaussian blur kernel is set to $\sigma  \sim \left[ {0.4,2} \right]$
\par During modeling training, we firstly sample 32 blurry LR video patches with a size of $64\times64$ in each mini-batch to train ${{\rm{{\cal N}}}_k}$ by optimizing ${{\rm{{\cal L}}}_{kernel}}$ with a learning rate of $1\times10^{-4}$. Subsequently, we train the entire model using the overall loss ${\rm{{\cal L}}} = {{\rm{{\cal L}}}_{kernel}} + {{\rm{{\cal L}}}_{SR}}$, where ${{\rm{{\cal L}}}_{SR}}$ is the pixel-wise difference ${{\rm{{\cal L}}}_{SR}} = {\left\| {I_t^{GT} - {I_t}^{SR}} \right\|_1}$ between restored mid-frame $I^{SR}_t$ and ground-truth $I^{GT}_t$ and $\rm{{\cal L}}_{kernel}$ is shown in Fig. \ref{fig3}(a). The learning rate is initialized to $1\times10^{-4}$ and halved for every 25 epochs, and finally, our BSVSR model will reach convergence after 50 epochs. Date augmentation \cite{zhang2020_PSTCR} is performed by random flipping and 90° rotation. We use the Adam optimizer for model optimization, and all experiments are conducted on a single NVIDIA RTX 3090 GPU with 24GB memory under the PyTorch framework.

\subsection{Comparison With State-of-the-Arts}
In this subsection, we comprehensively evaluate the effectiveness of our proposed BSVSR on synthetic and real-world satellite videos and compare it with state-of-the-art approaches. Both quantitative and qualitative results are presented and analyzed in detail.
\subsubsection{Selected Methods and Metrics}
The comparative approaches can be categorized into four types:
\begin{enumerate}[label=(\roman*)]
\item non-blind SISR model, as SwinIR \cite{liang2021swinir};
\item blind SISR methods, including IKC \cite{gu2019blind}, AdaTarget \cite{ada} and DASR \cite{wang2021unsupervised};
\item non-blind VSR networks, such as DUF-52L \cite{jo2018deep}, EDVR-L \cite{wang2019edvr}, BasicVSR \cite{chan2021basicvsr}, MSDTGP \cite{9530280} and MANA \cite{yu2022memory};
\item blind VSR approach DBVSR \cite{pan2021deep}.
\end{enumerate}
\par In particular, for blind SISR: IKC estimate the explicit blur kernels and transforms the blur information with affine transformation, termed Spatial Feature Transformation (SFT) layer. Now, the SFT layer is prevailing in blind SISR models; DASR predicts the degradation representation without kernel correction, which is more efficient than IKC. Besides, a Degradation-aware Convolution (DAConv) is proposed in DASR for effective blur transformation; AdaTarget is an implicit blur transformation model which does not require blur estimation. Therefore, we include these representative blind SISR methods for comparison. For non-blind VSR: DUF-52L implicitly realizes temporal compensation using 3D dynamic filters; EDVR-L and MSDTGP employ deformable convolution to explore temporal priors. Notably, MSDTGP is specialized for satellite videos; BasicVSR is a bi-directional propagation approach that provides leading performance with optical flow compensation; MANA relies on non-local attention for compensation. Hence, these methods provide comprehensive coverage of mainstream temporal compensation strategies. As for blind VSR, few studies are available for comparison. DBVSR combines optical flow warping and SFT layer to achieve blind VSR, which produces promising results against non-blind models.
\par Three metrics are involved in the quantitative evaluation. In experiments on simulated data, the classical image quality indicator Peak Signal-to-Noise Ratio (PSNR) and Structural Similarity Index Measure (SSIM) are used to measure the fidelity between restored results and ground truth. In real-world experiments, the reference-free metric Natural Image Quality Evaluator (NIQE) \cite{6353522} is adopted. NIQE compares the restored image to a default model computed from images of natural scenes. Note that a lower NIQE value indicates better quality of human perception.

\subsubsection{Quantitative Comparison}
Ascribing to the vital implementation of pixel-wise blur level measurement, our BSVSR is more flexible and allows accurate temporal compensation in blurry LR satellite videos. As reported in Table. \ref{table1}, \ref{tab2} and \ref{tab3}, BSVSR achieves the best average performances on all test sets and is well-generalized in various blur kernels with kernel width $\sigma=1.2, 1.6, 2.0$. 
\par For instance, when $\sigma=2.0$ on Jilin-1 testset in Table. \ref{table1}, our BSVSR gains a PSNR improvement of 4.21dB and 2.28dB compared to the best non-blind models SwinIR and EDVR-L, respectively. This illustrates that the classical SR approach is not capable of handling blind degradation processes, such as unknown blurs and downsampling. Compared to the best blind SISR method DASR, we can lead it by a large margin of 1.31dB, indicating that accurate temporal compensation is crucial for blind satellite VSR as it provides cleaner and sharper temporal cues to make the highly ill-posed problem well-posed. Although BSVSR shares the same blur estimation network with DBVSR, we can surpass it by 0.86dB, which demonstrates that temporal compensation also plays an important role in blind VSR. As mentioned before, most blind SR methods focus on blur estimation and overlook the critical aspect: temporal compensation. Under the same blur estimation procedure, the proposed progressive compensation strategy can extract more vital sharpness than the optical flow used in DBVSR, thus eliminating the interference of blurry pixels for better restoration.
\par Furthermore, as listed in Table \ref{tab2} and \ref{tab3}, we found that the blind SISR methods without temporal compensation can outperform the blind VSR approaches, such as IKC can ahead DBVSR 0.34dB in UrtheCast when $\sigma=1.2$. On the one hand, this illustrates that temporal compensation in severely blurred and downsampled videos remains a challenging task because instead of exploring more complementary, inaccurate temporal compensation may introduce interference that is harmful to restoration. On the other hand, we can alleviate this misalignment by performing pixel-wise blur-level modeling with deformable attention.

\subsubsection{Qualitative Comparison}
The visual comparison results are shown in Fig. \ref{jilin}, \ref{uc}, and \ref{sky}, the proposed method is capable of recovering more sharp and reliable details in various blurs and scenes. 
\par For example, as shown in Fig. \ref{jilin}, we observed that IKC and DBVSR restore apparent artifacts in Scene-2 of Jilin-1. This could be attributed to the mismatch between the estimated blur kernel or undesirable feature adaptation. Notably, the proposed BSVSR shares the same blur estimation component with DBVSR. However, it achieves favorable visual results, suggesting that the issue may not lie in the blur estimation but in the inaccurate blur transformation involved in DBVSR. Therefore, we can conclude that our pyramid spatial transformation can better calibrate the mid-feature to the correct solution space compared to SFT.
\par In Fig. \ref{uc}, AdaTarget, IKC, BasicVSR, and MANA generate severe line distortion and blending, which completely deviate from the ground truth. Our BSVSR reconstructs the spatial details and structures better than DBVSR, which illustrates deformable attention success in exploring more sharp details from blurred LR frames than inaccurate optical flow warping. Similarly, as shown in Fig. \ref{sky}, all of the comparative methods tend to recover more bent shapes and fuzzy edges of the buildings on the ground. Our BSVSR produces higher fidelity with reliable and sharp distributions against other non-blind and blind SR approaches.

\subsubsection{Real-world Comparison}
Apart from the experiments on simulated degradation, we further evaluate our method against state-of-the-art models on real-world satellite videos. The visual comparisons and corresponding NIQE metrics are shown in Fig. \ref{real}. Apparently, our method achieves the best NIQE, indicating the best human perception. Visually, IKC restores blur and noise on the edge of the plane, MANA creates unpleasing distortion in the tail, and our results are more clean and sharp in the high-frequency details.

\subsection{Ablation Studies}
In this section, we investigate the effectiveness of the key components of our BSVSR. Also, hyper-parameter setting and model efficiency are carefully discussed. 
\subsubsection{Effectiveness of Blur Kernel Estimation}
To examine the effects of blur kernel estimation, we build three models for comparison: a) Model-1: we remove the network ${{\rm{{\cal N}}}_k}$, ${{\rm{{\cal N}}}_{FFT}}$ and ${{\rm{{\cal N}}}_s}$. In this case, we pass $I^{LR}_t$ to five $3\times3$ convolution layers to generate ${\rm{{\cal F}}}_t^s$ for transformation; b) Model-2: we delete ${{\rm{{\cal N}}}_k}$ and do not estimate blur kernel $\rm{B_t}$. Here, we use bilinear upsampling to generate the latent sharp frame $I^{sharp}$; c) Model-3: we retain ${{\rm{{\cal N}}}_k}$ but do not optimize it with ${{\rm{{\cal L}}}_{kernel}}$. As reported in Table. \ref{ab1}, Model-1 gets the worse performance as bilinear upsampling could not incorporate the blur information for blur-aware transformation. Model-2 could find some sharp cues with the help of sharp features extraction network. However, without the guidance of $\rm{B_t}$, it is hard to adjust the solution space to an accurate domain. Although Model-3 performs blur kernel estimation, it can not find the accurate blur distribution without the ${{\rm{{\cal L}}}_{kernel}}$. By applying the entire blur estimation process, our method gains substantial improvements for blind VSR tasks, such as leading Model-2 and Model-3 by 0.28dB and 3.8dB in terms of PSNR, respectively.
\begin{table}[t]
  \centering
  \captionsetup{font={scriptsize}}
  \caption{Ablation studies of blur kernel estimation. The PSNR (dB) is calculated in Scene-2 of Jinlin-T with $\sigma=1.6$.}
\scalebox{0.9}{
    \begin{tabular}{ccccc}
\toprule
    Method & Model-1 & Model-2 (w/o ${{\mathop{\rm B}\nolimits} _t}$) & Model-3 (w/o ${{\rm{{\cal L}}}_{kernel}}$) & \textbf{Ours} \\
\midrule
    PSNR  &   31.58    &   31.47    &    27.95  &\textbf{31.75}  \\
    SSIM  &   0.9332    &    0.9278   &   0.8890    &\textbf{0.9361} \\
\bottomrule
    \end{tabular}%
}
  \label{ab1}%
\end{table}%

\begin{figure}[!t]
\centering
\includegraphics[width=3.5in]{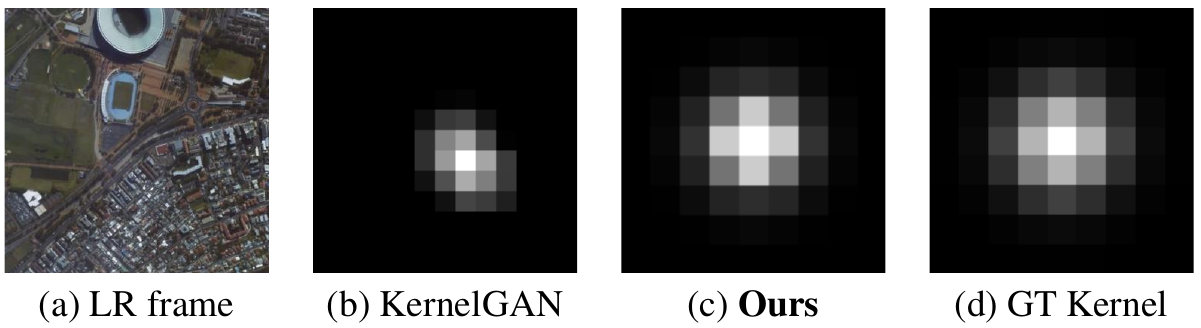}%
\captionsetup{font={scriptsize}}   
\caption{Visualization of estimated blur kernels. (a) Blurry low-resolution frame; (b) The estimated blur kernel of KernelGAN \cite{kgan}; (c) The estimated blur kernel of our BSVSR. (d) The ground-truth blur kernel.}
\label{kernel}
\end{figure}

\begin{table}[t]
  \centering
  \captionsetup{font={scriptsize}}
  \caption{Quantitative discussion of the progressive temporal compensation. One-stage means we directly align and fuse the $2N$ adjacent frames to the min-frame for temporal compensation. The PSNR (dB) is calculated in Scene-2 of Jinlin-T with $\sigma=1.6$.}
    \begin{tabular}{cccccc}
\toprule
    Fusion strategy & Models & Flow & MSD   & DA    & PSNR (dB) \\
\midrule
    \multirow{3}[0]{*}{One-stage} & Model-4 &  $\checkmark$     &       &       &31.44  \\
          & Model-5 &       &   $\checkmark$    &       &31.61  \\
          & Model-6 &       &       &    $\checkmark$   &31.68  \\
\midrule
    Progressive & \textbf{Ours}  &       &   $\checkmark$    &    $\checkmark$   &\textbf{31.75}  \\
\bottomrule
    \end{tabular}%
  \label{ab2}%
\end{table}%

\par Furthermore, as illustrated in Fig. \ref{kernel}, we visualize the estimated blur kernel $\rm{B_t}$ and compare it with another state-of-the-art blur kernel estimation approach KernelGAN \cite{kgan}. Our estimated blur kernel is consistent with the ground-truth kernel, which illustrates that our model predicts the precise blur information from satellite videos to make the network better aware of the degradation process.
\subsubsection{Effectiveness of Progressive Temporal Compensation}
As mentioned before, since the optical flow estimation is laborious and inaccurate to describe the motion relationship in severely blurry and low-resolution satellite videos, we introduce the Multi-Scale Deformable (MSD) convolution and Deformable Attention (DA) to progressively aggregate more clean and sharp cues in a coarse-to-fine manner. 
\par To begin with, we set up three baselines that directly align and fuse the blurry LR frames to generate the sharp mid-feature ${{\rm{\tilde {\cal F}}}_t}$ in a one-stage manner. Specifically, model-4 employs PWC-Net to estimate the optical flow maps from mid-frame $I^{LR}_t$ to adjacent frames. Then the warped adjacent features are concatenated and fused to ${{\rm{\tilde {\cal F}}}_t}$. Model-5 adopt MSD to align adjacent feature to min-feature, and the aligned features are concatenated and fused. Model-6 exploit DA to aggregate the frames. As listed in Table. \ref{ab2}, we find that progressive compensation shows more promising performance than one-stage compensation. As examined in previous works \cite{isobe2020video, 9530280}, dividing the entire frame sequences into sub-sequence helps to alleviate the alignment difficulties caused by large-displacement. Our method uses a slid-window strategy to progressively merge the temporal information in each window, relieving the pressure of MSD alignment. Besides, the optical-flow-based compensation model produces the worse results in one-stage compensation models. Therefore, we argue that optical flows in not robust and accurate in blurry LR satellite videos.
\par In addition, we set different combinations in the coarse-to-fine compensation process. The results are reported in Table. \ref{ab3}, which indicates that MSD is more practical to explore coarse sharpness and DA can aggregate the final sharp mid-features by considering the blur level of pixels.

\begin{figure}[!t]
\centering
\includegraphics[width=3in]{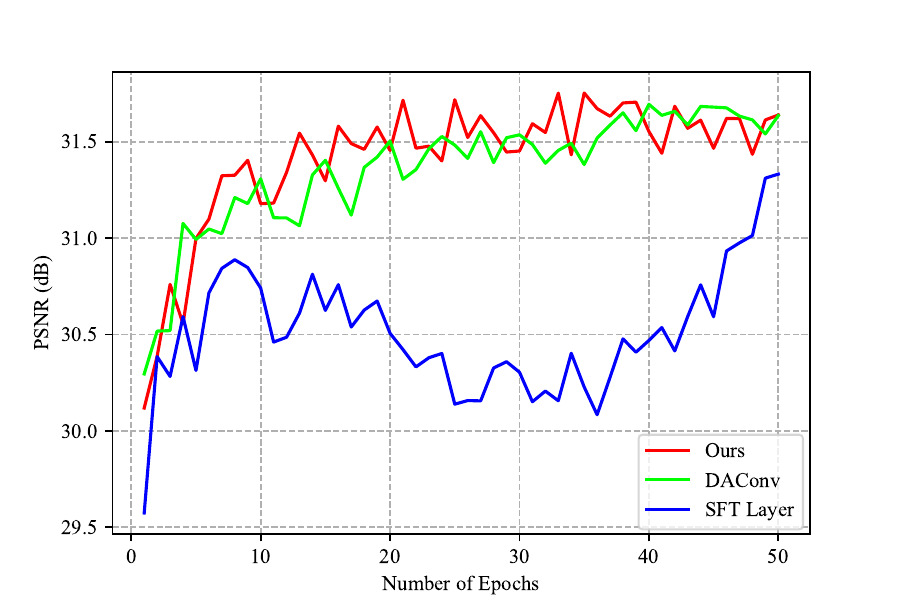}%
\captionsetup{font={scriptsize}}   
\caption{The training process of different transformation modules. Our Pyramid Spatial Transformation (PST) provides a more robust and effective blur transformation performance.}
\label{modulation}
\end{figure}

\begin{table}[!t]
  \centering
  \captionsetup{font={scriptsize}}
  \caption{Quantitative discussion of the coarse-to-fine temporal compensation. Model-7 to Model-11 adopt a different combination of fusion modules. The PSNR (dB) is calculated in Scene-2 of Jinlin-1 with $\sigma=1.6$.}
    \begin{tabular}{cccc}
\toprule
    Model & Fusion-1 (Coarse) & Fusion-2 (Fine)  & PSNR (dB) \\
\midrule
    Model-7 & Flow Warp + Conv  & ${{\rm{{\cal N}}}_{MSD}}$   &    31.58    \\
    Model-8 & Flow Warp + Conv  & ${{\rm{{\cal N}}}_{DA}}$    &   31.65      \\
    Model-9 & ${{\rm{{\cal N}}}_{DA}}$    & ${{\rm{{\cal N}}}_{MSD}}$   &  31.69       \\
    Model-10 & ${{\rm{{\cal N}}}_{DA}}$    & Flow Warp + Conv  &  31.54      \\
    Model-11 & ${{\rm{{\cal N}}}_{MSD}}$   & Flow Warp + Conv  &  31.42      \\
    \textbf{Ours}  & ${{\rm{{\cal N}}}_{MSD}}$   & ${{\rm{{\cal N}}}_{DA}}$    &  \textbf{31.75}       \\
\bottomrule
    \end{tabular}%

  \label{ab3}%
\end{table}%

\begin{table}[!t]
  \centering
  \captionsetup{font={scriptsize}}
  \caption{Ablation analysis on the effectiveness of Pyramid Spatial Transformation (PST). Baseline means we replace PST with three convolution layers. SFT and DA Conv are two widely used blur transformation approaches. The PSNR (dB) is calculated in Scene-2 of Jinlin-1 with $\sigma=1.6$.}
    \begin{tabular}{ccccc}
\toprule
    Model & Baseline & SFT Layer\cite{gu2019blind} & DA Conv \cite{wang2021unsupervised} & \textbf{PST (ours)} \\
\midrule
    PSNR  & 31.21&    31.33   &  31.69     & \textbf{31.75}  \\
    SSIM  &0.9287 &     0.9308  & 0.9345      &\textbf{0.9361}  \\
\bottomrule
    \end{tabular}%
  \label{ab4}%
\end{table}%

\begin{figure}[t]
\centering
\includegraphics[width=3.5in]{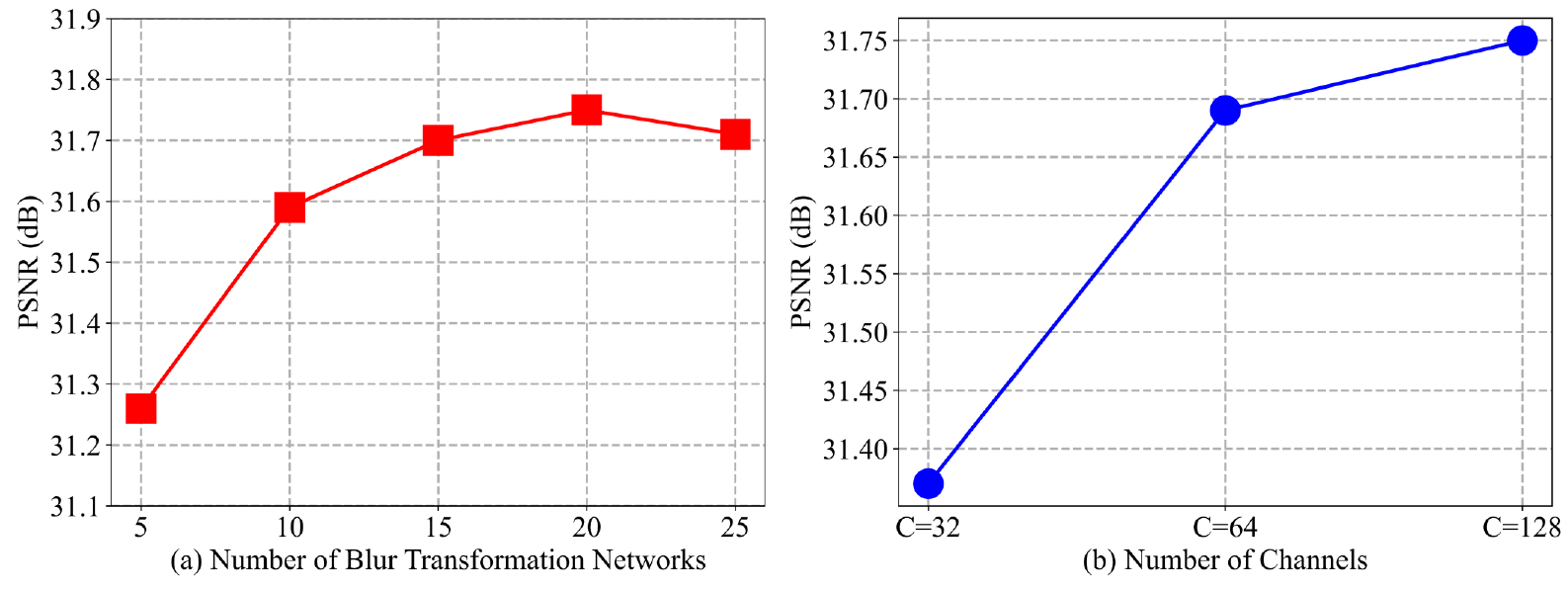}%
\captionsetup{font={scriptsize}}   
\caption{The impact of different hyper-parameters on the performance of our BSVSR.}
\label{num}
\end{figure}

\begin{table}[!t]
  \centering
  \captionsetup{font={scriptsize}}
  \caption{Quantitative comparisons of our method and SOTAs in terms of parameters, FLOPs, and running times. The FLOPs are calculated on 5 LR input frames with the size of $150\times150$. The running time (f/ps) is the average inference consumption per frame. Note that 1M=$10^6$ and 1G=$10^9$. Here, the PSNR (dB) is the average test result on five satellite videos.}
    \begin{tabular}{ccccc}
\toprule
    Methods & \#Param (M) & FLOPs (G)& Time (s)& PSNR (dB) \\
\midrule
    SwinIR \cite{liang2021swinir} & 11.8  & 266.8 &   0.054    &28.50  \\
\midrule
    IKC \cite{gu2019blind}   & 9.5   & 130.6 &   0.045    &31.78  \\
    AdaTarget \cite{ada} & 16.7  & 403.4 &    0.137   &28.56  \\
    DASR \cite{wang2021unsupervised}  & \textbf{5.6}   & \textbf{107.2} &   \textbf{0.022}    &32.12  \\
\midrule
    DUF-52L \cite{jo2018deep} & 6.8   & 736.6 &   0.078    &28.31  \\
    EDVR-L \cite{wang2019edvr} & 20.7  & 897.8 &      0.041 &30.57  \\
    BasicVSR \cite{chan2021basicvsr} & 6.3   & 163.7 &   0.034    &29.58  \\
    MSDTGP \cite{9530280} & 14.1  & 1579.8 &     0.128  &29.87  \\
    MANA \cite{yu2022memory}  & 22.2  & 633.5 &    0.056   & 29.32 \\
\midrule
    DBVSR \cite{pan2021deep} & 14.1 & 1792.3 &    0.154   & 32.20 \\
    \textbf{BSVSR (ours)}  & 15.2  & 688.9 &    0.077   &\textbf{32.95}  \\
\bottomrule
    \end{tabular}%
  \label{ab5}%
\end{table}%

\begin{figure}[!b]
\centering
\includegraphics[width=3in]{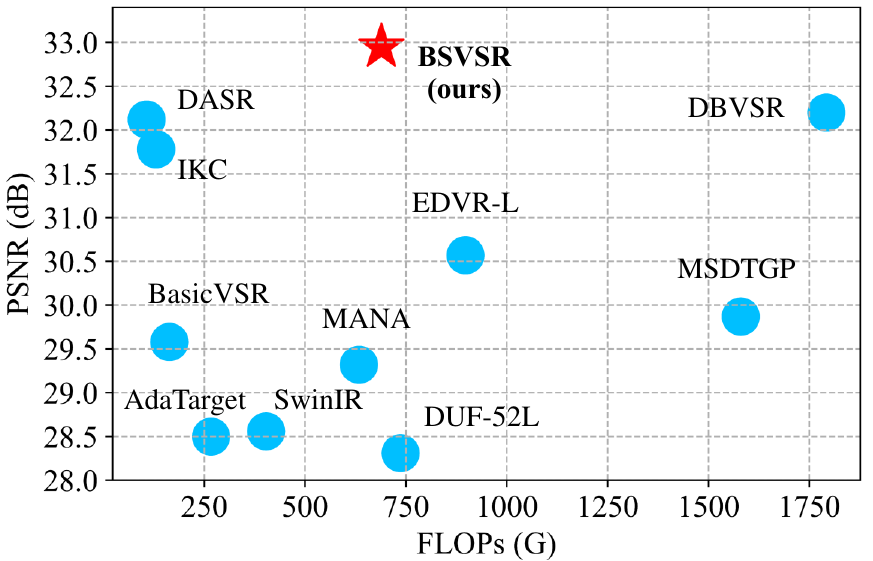}%
\captionsetup{font={scriptsize}}   
\caption{The relationship between FLOPs (G) and PSNR (dB) performance. Our BSVSR achieves a favorable trade-off between computational complexity and performance.}
\label{flops}
\end{figure}

\subsubsection{Effectiveness of Pyramid Spatial Transformation}
With the awareness of blur information, we perform flexible transform to adjust the mid-feature in the spatial and channel dimension to make it adaptive to various degradations. To evaluate the proposed Pyramid Spatial Transformation (PST), we set a baseline model by replacing PST with three convolution layers. In addition, two widely used blur transformation modules, Spatial Feature Transformation (SFT) layer and Degradation-Aware Convolution (DAConv), are introduced as alternative transformation networks for comparison. The training process is displayed in Fig. \ref{modulation}, and the quantitative results can be found in Table. \ref{ab4}. We observe that the SFT layer suffers instability in the training process. Benefiting from the well-defined spatial and channel activation, our method reflects robust blur transformation. Moreover, our PST ahead SFT layer and DA Conv of PNSR by 0.42dB and 0.06dB, respectively. As mentioned before, the pyramid structure is beneficial for preserving the multi-scale information in satellite videos. Therefore, we could conduct multi-level transformation to ensure more precise feature adaptation.

\subsubsection{Model Efficiency}
Here we firstly discuss the relationship between model size and performance of BSVSR by stacking different numbers $n$ of transformation network ${{\rm{{\cal N}}}_{n}^T}$. The quantitative of PSNR is provided in Fig. \ref{num} (a). With the growing number of $n$, the communication between blur information and min-feature deepens, leading to better feature adaptation. However, such improvement reaches a plateau when $n=15$. Additionally, we investigate the impact of channel number $C$. As shown in Fig. \ref{num} (b), despite increasing performance as $C$ grows, the huge parameters of our model become an urgent issue when $C$ is larger than 64. To achieve a favorable trade-off between model efficiency and performance, we adopt $n=20$ and $C=128$ in our final model.
\par Furthermore, we evaluate the model efficiency of comparative state-of-the-art methods in terms of parameters, FLOating Point operations (FLOPs), and running times. As reported in Table. \ref{ab5}, BSVSR gains the best performance with an acceptable model size. Note that SISR methods naturally have fewer parameters and FLOPs than the VSR approach as they do not include extra components for temporal compensation. Compared to the best non-blind VSR model EDVR-L, BSVSR is 27\% less in parameters (15.2M vs. 20.7M) and 208.9 less in FLOPs. Regarding the blind VSR approach DBVSR, our BSVSR has significantly lower complexities (688.9G vs. 1792.3G) and faster inference speed (0.077s vs. 0.154s). This is because the optical flow estimation used in DBVSR is time-consuming, whereas our progressive compensation strategy is more computationally efficient and effective in temporal compensation. Fig. \ref{flops} gives a visual relationship between model performance and FLOPs. It is evident that our method achieved favorable performance compared to other methods while maintaining affordable complexity.

\section{Conclusion}\label{conclu}
In this paper, we proposed a blind SR network for satellite videos (BSVSR). Our key motivation is that not all the pixels can provide clean and sharp cues for blind VSR. Therefore, our BSVSR mainly aims at compensating for blurry and smooth pixels from severely degraded satellite videos by considering the blur level of pixels. Unlike  prior blind satellite VSR methods that employ optical flow warping or patch-wise similarity to align frames, we progressively explore temporal redundancy using Multi-Scale Deformable (MSD) convolution and further aggregate them into a sharp mid-feature with multi-scale Deformable Attention (DA) in a coarse-to-fine manner. Additionally, we devise a robust pyramid spatial transformation module, which recalibrates the sharp mid-feature in the multi-level domain to modulate the mid-feature into a suitable solution space. Extensive experiments on five video satellites demonstrate our BSVSR gains favorable performance against state-of-the-art non-blind and blind SR approaches.

Although the proposed BSVSR could explore vital sharpness from blurry LR satellite videos, it still remains a challenging task to grasp beneficial information in compressed satellite imagery. Besides, missing a real-world satellite video dataset poses a domain gap between simulation and reality. In our future work, we plan to take the noise term into consideration and build a large-scale dataset with real-world degradations.



%
%



%

%


%
%

\ifCLASSOPTIONcaptionsoff
  \newpage
\fi



%
\bibliographystyle{IEEEtran}
\bibliography{reference}


%

\begin{IEEEbiography}[{\includegraphics[width=1in,height=1.25in,clip,keepaspectratio]{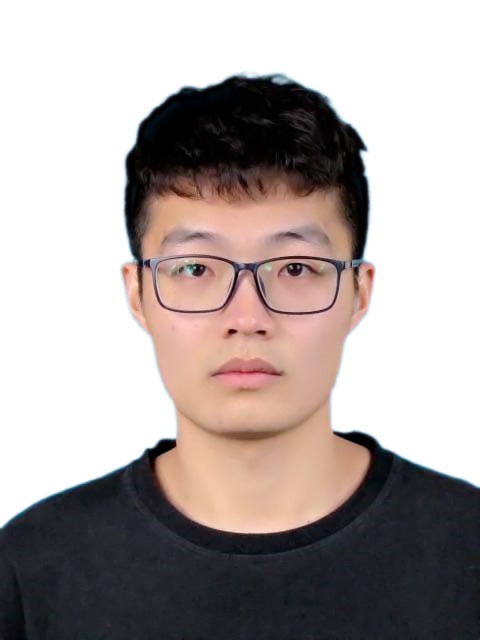}}]{Yi Xiao}
received the B.S. degree from the School of Mathematics and Physics, China University of Geosciences, Wuhan, China, in 2020. He is pursuing the Ph.D. degree with the School of Geodesy and Geomatics, Wuhan University, Wuhan.
\par His major research interests are remote sensing image super-resolution and computer vision. More details can be found at \href{https://xy-boy.github.io}{\textcolor{black}{https://xy-boy.github.io}}.
\end{IEEEbiography}

\begin{IEEEbiography}[{\includegraphics[width=1in,height=1.25in,clip,keepaspectratio]{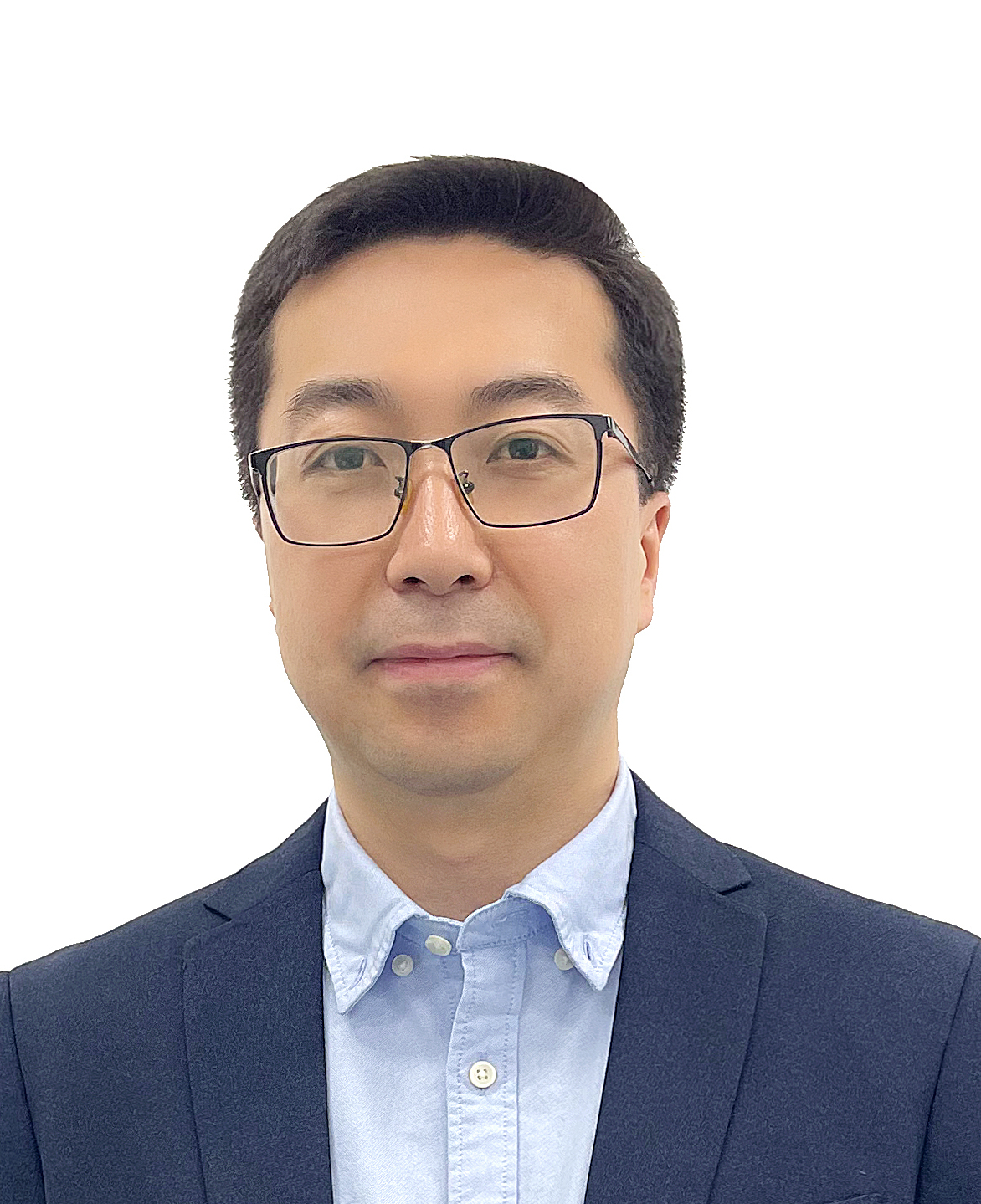}}]{Qiangqiang Yuan}
(Member, IEEE) received the B.S. degree in surveying and mapping engineering and the Ph.D. degree in photogrammetry and remote sensing from Wuhan University, Wuhan, China, in 2006 and 2012, respectively.
\par In 2012, he joined the School of Geodesy and Geomatics, Wuhan University, where he is a Professor. He has published more than 90 research papers, including more than 70 peer-reviewed articles in international journals, such as \emph{Remote Sensing of Environment, ISPRS Journal of Photogrammetry and Remote Sensing}, {\sc IEEE Transaction ON Image Processing}, and {\sc IEEE Transactions ON Geoscience AND Remote Sensing}. His research interests include image reconstruction, remote sensing image processing and application, and data fusion.
\par Dr. Yuan was a recipient of the Youth Talent Support Program of China in 2019, the Top-Ten Academic Star of Wuhan University in 2011, and the recognition of Best Reviewers of the IEEE GRSL in 2019. In 2014, he received the Hong Kong Scholar Award from the Society of Hong Kong Scholars and the China National Postdoctoral Council. He is an associate editor of 5 international journals and has frequently served as a referee for more than 40 international journals for remote sensing and image processing.
\end{IEEEbiography}

\begin{IEEEbiography}[{\includegraphics[width=1in,height=1.25in,clip,keepaspectratio]{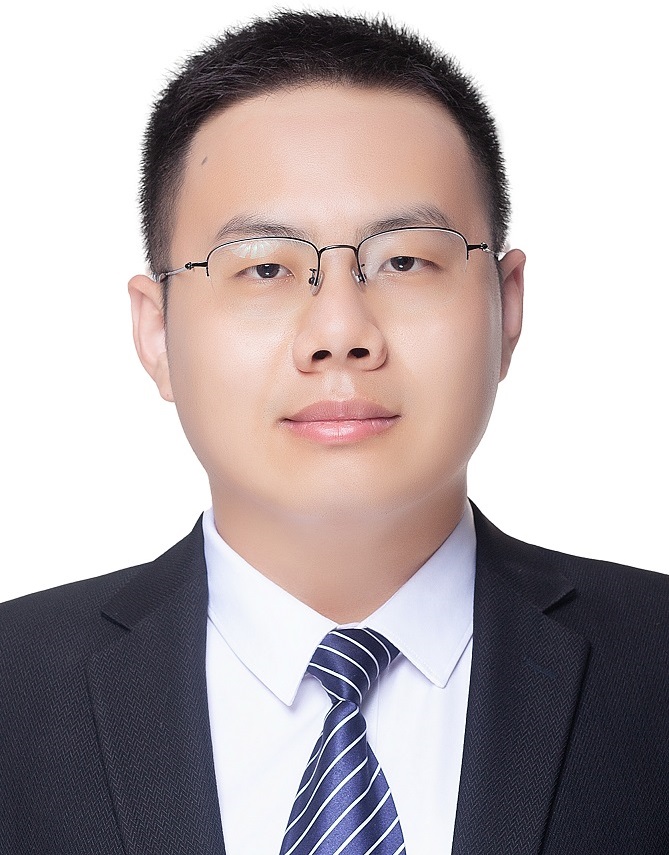}}]{Qiang Zhang}
received the B.E. degree in surveying and mapping engineering, M.E. and Ph.D. degree in photogrammetry and remote sensing from Wuhan University, Wuhan, China, in 2017, 2019 and 2022, respectively.
	\par He is currently an Associate Professor with the Center of Hyperspectral Imaging in Remote Sensing (CHIRS), Information Science and Technology College, Dalian Maritime University. His research interests include  remote sensing information processing, computer vision, and machine learning. He has published more than ten journal papers on IEEE TIP, IEEE TGRS, ESSD, and ISPRS P\&RS. More details could be found at \href{https://qzhang95.github.io}{\textcolor{black}{https://qzhang95.github.io}}.
\end{IEEEbiography}



\begin{IEEEbiography}[{\includegraphics[width=1in,height=1.25in,clip,keepaspectratio]{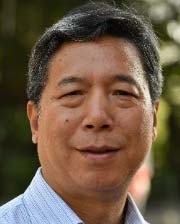}}]{Liangpei Zhang}
(Fellow, IEEE) received the B.S. degree in physics from Hunan Normal University, Changsha, China, in 1982, the M.S. degree in optics from the Xi’an Institute of Optics and Precision Mechanics, Chinese Academy of Sciences, Xi’an, China, in 1988, and the Ph.D. degree in photogrammetry and remote sensing from Wuhan University, Wuhan, China, in 1998.
\par He is currently a “Chang-Jiang Scholar” Chair Professor appointed by the Ministry of Education of China at the State Key Laboratory of Information Engineering in Surveying, Mapping, and Remote Sensing (LIESMARS), Wuhan University. He was a Principal Scientist for the China State Key Basic Research Project from 2011 to 2016 appointed by the Ministry of National Science and Technology of China to lead the Remote Sensing Program in China. He has published more than 700 research articles and five books. He is the Institute for Scientific Information (ISI) Highly Cited Author. He holds 30 patents. His research interests include hyperspectral remote sensing, high-resolution remote sensing, image processing, and artificial intelligence. 
\par Dr. Zhang is a fellow of the Institution of Engineering and Technology (IET). He was a recipient of the 2010 Best Paper Boeing Award, the 2013 Best Paper ERDAS Award from the American Society of Photogrammetry and Remote Sensing (ASPRS), and the 2016 Best Paper Theoretical Innovation Award from the International Society for Optics and Photonics (SPIE). His research teams won the top three prizes in the IEEE GRSS 2014 Data Fusion Contest. His students have been selected as the winners or finalists of the IEEE International Geoscience and Remote Sensing Symposium (IGARSS) Student Paper Contest in recent years. He is also the Founding Chair of the IEEE Geoscience and Remote Sensing Society (GRSS) Wuhan Chapter. He also serves as an associate editor or editor for more than ten international journals. He is also serving as an Associate Editor for the {\sc IEEE Transactions ON Geoscience AND Remote Sensing}.
\end{IEEEbiography}

\end{document}